\documentclass[journal]{IEEEtran}
\ifCLASSINFOpdf
\else
\fi
\usepackage{bm}
\usepackage{amsmath}
\usepackage{graphicx}
\usepackage{xcolor} 
\usepackage[colorlinks=true, linkcolor=blue,citecolor=blue,urlcolor=black]{hyperref} 
\usepackage{amssymb} 
\usepackage[ruled,vlined,linesnumbered]{algorithm2e} 

\usepackage{multirow}
\usepackage{tabularray}
\usepackage{booktabs}
\usepackage{eurosym}
\usepackage{makecell}
\usepackage[export]{adjustbox}
\usepackage{xcolor}
\UseTblrLibrary{booktabs}
\usepackage{float}
\newcommand{\ms}[2]{{\scriptsize #1\raisebox{-0.45ex}{\tiny$\pm$#2}}}

\begin{document}
%
\title{HALight: Heterogeneous Multi-Agent Reinforcement Learning for Traffic Signal Control under Lane-Level Missing Observations}
%
%
%

\author{Chenkai~Hu, Yingqian~Zhang, Ming~Xie, Yi~Fang, Chengxiang~Jin, Chao~Li,~\IEEEmembership{Member,~IEEE}
\thanks{Chenkai Hu, Ming Xie, Yi Fang and Chao Li are with the Department of Control Science and Engineering, Zhejiang University, Hangzhou 310027, China (e-mail: chenkaihu@zju.edu.cn; mingxie@zju.edu.cn; 12232096@zju.edu.cn; chaoli@zju.edu.cn).}
\thanks{Yingqian Zhang is with Zhejiang Institute of Communications Co.,LTD., Hangzhou, Zhejiang, 310030, China,  and also with the College of Control Science and Engineering, Zhejiang University, Hangzhou, Zhejiang, 310027, China (e-mail: zhangyq@zuaa.zju.edu.cn).}
\thanks{Chengxiang Jin is with the Institute of Cyberspace Security, College of Information Engineering, Zhejiang University of Technology, Hangzhou 310023, China (e-mail: jincxiang@zjut.edu.cn).}}

%
%

\markboth{IEEE TRANSACTIONS ON INTELLIGENT TRANSPORTATION SYSTEMS}%
{HU \MakeLowercase{\textit{et al.}}: HALight: Heterogeneous Reinforcement Learning for Traffic Signal Control}
%



\maketitle

\begin{abstract}
Multi-agent reinforcement learning has shown great promise for traffic signal control. However, most existing methods rely on two key assumptions: (i) agents are assumed to be homogeneous with unified observation/action spaces; (ii) full and reliable lane-level traffic observations are available to all agents. In practice, urban road networks exhibit substantial structural heterogeneity (e.g., lane configurations and phase designs), while sensor failures and incomplete detector coverage often result in missing incoming lane measurements. Such issues can severely impair coordination and destabilize learning. To address these challenges, we propose HALight, a communication-enhanced heterogeneous multi-agent reinforcement learning framework for traffic signal control under lane-level missing observations. HALight supports intersection-specific observation and action modeling to accommodate heterogeneous intersections, and incorporates a lightweight one-hop neighborhood communication module to enrich spatial context when lane-level observations are partially unavailable. Moreover, HALight adopts a sequential policy update scheme to improve coordination among heterogeneous agents. Experimental results on two real-world road networks demonstrate that HALight outperforms representative homogeneous and heterogeneous baselines in terms of traffic efficiency and convergence behavior. The code and datasets used in this work are available at https://github.com/Huchenkaii/HALight.
\end{abstract}


%
\IEEEpeerreviewmaketitle

\section{Introduction}
\IEEEPARstart{T}{raffic}  signal control (TSC) plays a crucial role in reducing traffic congestion~\cite{keyvan2015multiple} and carbon emissions~\cite{wu2025big} by optimizing signal timing and phase switching. Traditional methods are generally model-driven, relying on parametric formulations to minimize vehicle delay and queue length~\cite{koonce2008traffic,yao2020dynamic}. These approaches typically depend on explicitly specified traffic models and handcrafted objective functions. For instance, MaxBand~\cite{little1981maxband} computes signal timings by optimizing progression bandwidth using predefined traffic parameters. While such methods have achieved success in controlled settings, their applicability becomes limited in urban road networks~\cite{gershenson2004self}. It is difficult to model the complex and highly dynamic traffic environment arising from interactions among multiple intersections. As a result, traditional approaches often struggle to adapt to time-varying multi-intersection conditions.

\begin{figure}[htbp]
    \centering
    \includegraphics[width=0.8\columnwidth]{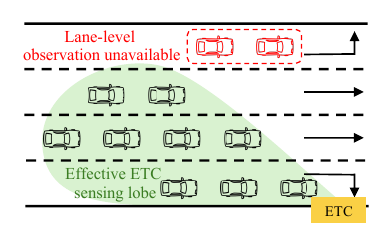}
    \caption{Illustration of lane-level observation unavailability under non-uniform ETC sensing coverage.}
    \label{fig:etc}
\end{figure}

With the rapid progress of reinforcement learning (RL) and deep neural networks (DNNs), deep reinforcement learning (DRL) has emerged as a data-driven paradigm for TSC~\cite{el2013multiagent,wei2018intellilight,zhang2022distributed}. Existing multi-agent reinforcement learning (MARL) methods directly learn control policies through trial-and-error interactions~\cite{chen2020toward, wu2020multi}. By assigning an individual learning agent to each intersection, decentralized approaches have become a popular solution for multi-intersection control. TeDA-GCRL~\cite{lin2023temporal} explicitly models lane-level interactions via graph convolution, where each lane is treated as a graph node to capture relations of intersections. CoordLight~\cite{zhang2025coordlight} introduces Queue Dynamic State Encoding (QDSE) to characterize queue dynamics. Despite their success, decentralized MARL methods for TSC are constrained by two key assumptions that limit their applicability in real-world road networks. (i) Most methods assume homogeneous agents with shared observation and action spaces across intersections~\cite{wei2019colight,gu2024large}. While this simplification aids learning and coordination, it fails to account for the structural differences in intersection geometry, lane configuration, and phase design. As illustrated in Fig.~\ref{fig:intro}, we identify two common types of heterogeneity: the variation across different intersection types and the structural heterogeneity within the same intersection type. Consequently, the learned policies tend to favor an ``average" behavior, which is suboptimal for intersections with distinct characteristics. (ii) Existing methods typically assume full lane-level observations~\cite{wei2019presslight,jiang2024general}. In practical deployments, sensor failures and incomplete or spatially varying sensing coverage can lead to missing incoming-lane measurements, which may persist over a control horizon or occur intermittently~\cite{liang2022memory}. For example, the effective communication coverage of electronic toll collection (ETC) roadside units (RSUs) is determined by the spatial field distribution of the RSU antenna and can vary across roadway locations and lanes~\cite{zhang2012communication}.
As illustrated in Fig.~\ref{fig:etc}, such non-uniform coverage may result in unavailable traffic observations over particular lane segments. The lack of reliable lane-level information limits agents' adaptation to local conditions. Combined with homogeneous agent modeling, it further degrades coordination performance.

Such limitations motivate the development of heterogeneous multi-agent TSC under lane-level missing observations. However, developing this model presents two primary challenges:

\begin{figure}[t]
    \centering
    \includegraphics[scale=0.6]{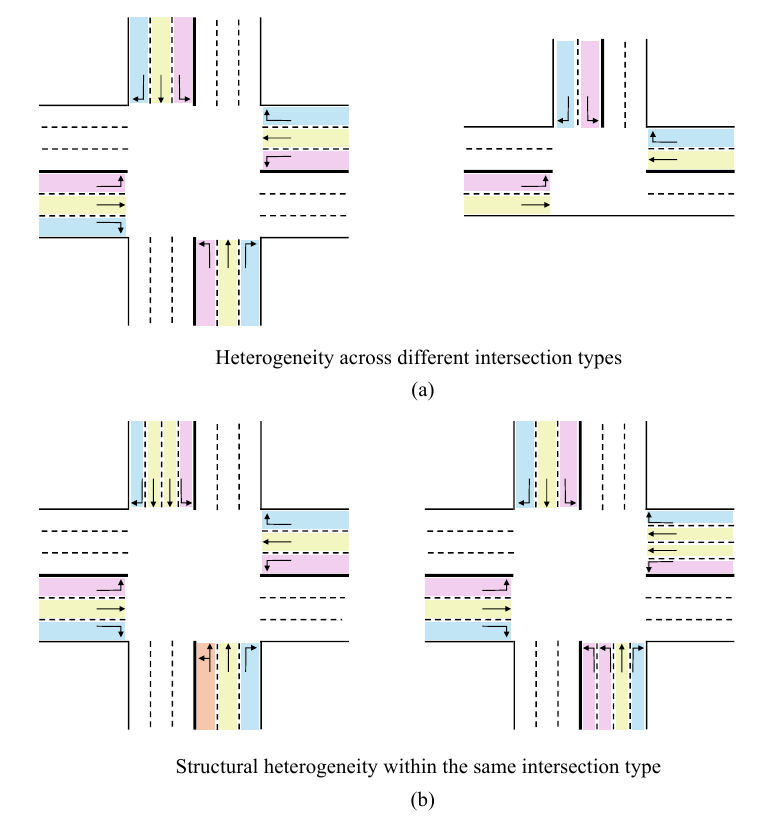}
    \caption{Illustration of two prevalent forms of intersection heterogeneity in real-world road networks: 
(a) heterogeneity across different intersection types, such as crossing intersections and T intersections,
(b) heterogeneity among intersections of the same type, characterized by variations in lane configurations, geometric layouts, and phase designs.
}
    \label{fig:intro}
\end{figure}

\begin{itemize}
\item How to model the observation and action spaces across heterogeneous intersections? In real-world road networks, there exist substantial variations in intersection types and configurations. The variations lead to significant differences in the observation and action spaces across agents. Most heterogeneous methods focus on the heterogeneity between crossing and T intersections~\cite{wang2024large, lu2025haps}. Therefore, these methods are restricted to specific types of road networks and fail to adapt to real-world conditions.

\item How to enable heterogeneous agents to remain effective under different lane-level observation missing patterns? Each intersection may experience unique missing observation patterns. Such missing patterns can vary both across intersections and within the same intersection. The diversity of missing patterns, combined with the heterogeneity between intersections and agents, complicates the learning process and makes it more challenging for the multi-agent system to adapt to varying local observations.
\end{itemize}

To solve these issues, we propose HALight, a heterogeneous MARL framework for TSC under lane-level missing observations. To address the first challenge, HALight formulates TSC with heterogeneous agents and adopts intersection-specific observation and action representations. HALight explicitly tailors each agent’s representation to its local lane configuration and phase design, enabling agents to account for structural differences among intersections and overcome the limitations of homogeneous agent modeling. For the second challenge, HALight incorporates a lightweight heterogeneous communication mechanism. It allows agents to selectively aggregate information from one-hop neighboring intersections, thereby mitigating the impact of locally missing observations and improving coordination. The main contributions are summarized as follows:

\begin{enumerate}
    \item We investigate a practically important problem: enabling effective decentralized MARL for TSC in heterogeneous road networks under lane-level missing observations. In real-world deployments, intersections differ in types and configurations, and lane-level traffic measurements are often incomplete. The coexistence of intersection heterogeneity and lane-level observation missingness poses significant challenges for decentralized MARL methods.

    \item We propose HALight, a communication-enhanced heterogeneous MARL framework for TSC. HALight formulates TSC with heterogeneous agents by adopting intersection-specific observation and action representations tailored to local lane configurations and phase designs. In addition, HALight incorporates a lightweight one-hop communication mechanism to alleviate the impact of locally missing lane-level observations and improve coordination in heterogeneous road networks.

    \item Extensive experiments are conducted on two real-world road networks in Hangzhou under varying  demands and observation missing rates. The results demonstrate that HALight outperforms representative baselines in terms of efficiency and convergence behavior.
\end{enumerate}

The rest of the paper is organized as follows. Section~\ref{sec:related} reviews related work. Section~\ref{sec:formulation} formulates the TSC problem under a MARL framework. Section~\ref{sec:HALight} presents the proposed method. Section~\ref{sec:experiments} describes the experimental settings and discusses the results. Finally, Section~\ref{sec:conclusion} concludes the paper.

\section{RELATED WORK}
\label{sec:related}
This section provides a summary of DRL-based multi-intersection cooperative control methods, covering both homogeneous and heterogeneous control strategies, and discussing lane-level missing observations in TSC problems.
\subsection{Homogeneous MARL for Traffic Signal Control}
Homogeneous MARL assumes that all agents share identical observation spaces, action spaces, and network architectures. Under this assumption, one commonly adopted early approach is the independent MARL paradigm. For example, \cite{van2016coordinated} modeled the traffic network as a collection of independently controlled intersections, with coordination achieved through transfer planning and the max-plus algorithm. PressLight~\cite{wei2019presslight} further incorporated the Max-pressure~\cite{varaiya2013max} principle into an IDQN framework, allowing agents to optimize the pressure difference between upstream and downstream intersections. However, independent MARL methods treat other agents as part of the environment, often resulting in unstable training as neighboring policies change.

To mitigate this issue, the centralized training and decentralized execution (CTDE) paradigm~\cite{lowe2017multi,son2019qtran} is often coupled with parameter sharing~\cite{gupta2017cooperative} in MARL-based TSC~\cite{jiang2024general,huang2021network}. In such shared policy implementations, agents are trained with centralized information while sharing policy parameters, which allows them to benefit from collective experiences and improves sample efficiency.  For instance, RegionLight~\cite{gu2024large} integrates CTDE with regional partitioning and an Adaptive Branching Dueling Q-Network (ABDQ) for dynamic action space decomposition, enabling coordinated decision making while maintaining decentralized execution. 
To better capture spatial dependencies among intersections, some MARL methods incorporate graph neural networks (GNNs) to aggregate information from neighboring intersections~\cite{wei2019colight,nishi2018traffic,devailly2021ig}. In addition, sequential decision-making has been considered in TSC~\cite{fan2025centralized,zhang2025learning}. For example, GNSD-Light~\cite{liu2025sequential} introduces a priority-aware sequential decision MARL framework to coordinate intersections with different priorities.



While homogeneous MARL methods have shown promising performance, they typically rely on unified observation/action representations. This reliance limits their applicability to heterogeneous road networks where intersections differ in lane configuration and phase set. Relative indicators, such as pressure, density and occupancy ratio, can partially alleviate observation space differences across intersections~\cite{zhang2025coordlight,wei2019presslight}. However, when intersections have different phase sets, constructing a unified action space through action masking can still lead agents to learn averaged behaviors, limiting their ability to learn intersection-specific control policies.

\subsection{Heterogeneous MARL for Traffic Signal Control}

Real-world traffic networks contain intersections with diverse lane configurations and signal phase sets, leading to different control requirements. Heterogeneous MARL therefore enables intersections to learn intersection-specific policies while coordinating with structurally different agents.

Several studies have explored heterogeneous MARL for TSC. Early work assigns independent agents with intersection-specific policies to heterogeneous intersections~\cite{calvo2018heterogeneous}. Other studies reduce structural differences through transformation or unified representations. HA2C~\cite{shen2024heterogeneous} employs an intersection structure transformation scheme and learns both general and structure-specific policies, while HeteroLight~\cite{zhang2024heterolight} introduces general feature extraction to construct unified state and action representations across heterogeneous intersections. More recent methods further focus on generalization across diverse traffic scenarios. CROSS~\cite{chen2026cross} employs a mixture-of-experts policy to enable adaptive specialization, whereas Unicorn~\cite{zhang2026unicorn} combines universal and intersection-specific representations for heterogeneous network control. These studies demonstrate different strategies for balancing shared representations with intersection-specific characteristics.

Beyond representation design, heterogeneous agent reinforcement learning has also studied policy optimization for agents with distinct policies. HAPPO and HATRPO~\cite{kuba2022trust} employ sequential policy updates based on the multi-agent advantage decomposition lemma~\cite{kuba2021settling}, providing theoretically justified monotonic policy improvement. Heterogeneous-Agent Mirror Learning ~\cite{zhong2024heterogeneous} further generalizes the theoretical framework for heterogeneous policy optimization. However, these theoretical analyses are established under the fully observable setting considered in their original formulations. When agents operate with incomplete observations and recurrent histories, the original monotonic-improvement argument does not directly apply. This motivates further study of heterogeneous policy optimization under partial observability in TSC.

\subsection{Lane-level Missing Observations in Traffic Signal Control}

Lane-level observations can become incomplete in real-world TSC because of sensor failures, limited sensing coverage, or sparse detector deployment~\cite{liang2022memory}. Existing traffic data imputation methods~\cite{han2020content,zhang2022self,chen2023novel} mainly focus on reconstructing incomplete traffic measurements. However, applying a separate reconstruction process to online TSC introduces additional processing before decision making.

To address partial observability directly within the decision-making process, recurrent models are commonly used to encode historical observations and retain temporal context~\cite{hausknecht2015deep,wang2024emarlin}. Such temporal information can help distinguish similar current observations associated with different underlying traffic conditions. Meanwhile, spatiotemporal graph-based TSC methods like DynSTGAT~\cite{wu2021dynstgat} jointly model historical traffic dynamics and dependencies among neighboring intersections to improve coordinated control.

Communication provides another way to supplement incomplete local information. In general heterogeneous MARL, the Communication Complementary Graph Model (CCGM)~\cite{peng2024enhancing} incorporates messages from other agents into local observations through dynamically constructed communication graphs. Such communication mechanisms demonstrate the potential of spatial information for mitigating limited observations. For TSC, where traffic interactions are largely driven by neighboring upstream and downstream intersections, efficiently combining historical context with localized inter-agent information is particularly important. This motivates a lightweight temporal-spatial modeling strategy for heterogeneous TSC under lane-level missing observations.

\section{PROBLEM FORMULATION}
\label{sec:formulation}
This section provides a formal definition of the TSC problem and a comprehensive overview of the partially observable reinforcement learning framework. Since intersections in practical road networks can exhibit diverse structural configurations, we formulate and discuss the problem separately under heterogeneous and homogeneous modeling settings, to clearly distinguish their observation, action, and reward designs.
\subsection{Traffic Signal Control Definition}
For heterogeneous methods, we model the traffic network as a directed heterogeneous graph $G_{\text{het}}=(V,E)$, where each node $v_i\in V$ corresponds to a signalized intersection, and each directed edge $e_{ij}\in E$ represents a road segment that allows vehicles to flow from intersection $v_i$  to $v_j$. The network is heterogeneous in the sense that each intersection has distinct structural properties, such as different numbers of lanes and signal phase configurations, leading to diverse observation and action spaces across agents. In this setting, each intersection is controlled by a heterogeneously modeled agent equipped with an individualized policy network that receives a local observation $o_t^i$  and selects a control phase $a_t^i$ from its predefined phase set $\mathcal A^i$ at every decision step.

For homogeneous methods, we model the traffic network as a directed homogeneous graph $G_{\text{homo}}=(V^{'},E^{'})$, which preserves the same road connectivity as $G_{\text{het}}$  but assumes identical structural and operation phases for all intersections. In this setting, the observation space $\mathcal O$ is obtained by aggregating the incoming lane features of each intersection into a fixed size feature vector following a unified aggregation rule, so that all agents share the same observation dimension. Similarly, the action space $\mathcal A$ is defined as the union of all intersection phase sets, representing a common set of feasible signal phases available to every agent. Each intersection is controlled by an agent parameterized through a shared policy network which receives an observation $o_t\in \mathcal O$ and outputs a phase selection $a_t\in \mathcal A$ at each decision step.

\subsection{Partially Observable MARL Formulation}
MARL provides a framework in which multiple agents learn optimal policies through trial-and-error interactions with the environment. In practical TSC deployments, sensor failures and incomplete or spatially varying sensing coverage can lead to missing lane-level traffic observations, preventing agents from directly observing the complete environment state. Such settings can be naturally formulated as a Decentralized Partially Observable Markov Decision Process (Dec-POMDP), defined by a tuple $\mathcal{M}=<\mathcal S,\mathcal N,\mathcal O,\mathcal A,P,r,\gamma>$. $s\in \mathcal S$ is the state of the environment, $i \in \mathcal N \equiv \{1,2,\dots,n\}$ denotes the index of agents. The joint action space $\mathcal{A}=\prod \limits_{i=1}^{n}\mathcal A^i$ is the Cartesian product of all agents’ action space. $\gamma$ is the discount factor that determines the relative importance of future rewards compared to immediate ones. At each timestep $t$, agent $i$ receives the local observation $o^i$ from the environment and selects an action $a_t^i$ from predefined action set $\mathcal A^i$ according to its policy $\pi^{i}\left(a_{t}^{i} \mid o_{t}^{i}\right)$. The joint action $\bm {a}_t=(a_t^1,a_t^2,\dots,a_t^{n})$ and the joint policy $\boldsymbol{\pi}\left(\bm {a}_t \mid o_{t}\right)=\prod \limits_{i=1}^{n} \pi^{i}\left(a_{t}^{i} \mid o_{t}^{i}\right)$ are obtained by combining all individual agents’ actions and policies, respectively. After all agents take action, each agent $i$ receives a local reward $r_t^i$ and the environment transitions to the next state $s_{t+1}$  according to the probability function $P:\mathcal S \times \mathcal A \times \mathcal S \rightarrow [0,1]$. The system-level reward $r_t=\sum \limits_{i=1}^{n}r_t^i$ is obtained by aggregating all local rewards, which is consistent with the cooperative Dec-POMDP formulation. The state value function in a state $s$  is defined as $V_{\pi}(s)=\mathbb{E}_{\pi}\left[\sum \limits_{t=0}^{\infty} \gamma^{t} r_{t} \mid s_{0}=s, \bm {a}_t \sim \boldsymbol{\pi}, s_{t+1} \sim P\right]$ , and the action value of a state-action pair is defined as $Q_{\pi}(s, \bm {a})=\mathbb{E}_{\pi}\left[\sum \limits_{t=0}^{\infty} \gamma^{t} r_{t} \mid s_{0}=s, \bm {a}_0=\bm {a}, \bm {a}_t \sim \boldsymbol{\pi}, s_{t+1} \sim P\right]$. The advantage function is then given by $A_{\pi}(s, \bm{a})=Q_{\pi}(s, \bm{a})-V_{\pi}(s)$, which measures the relative benefit of taking joint action  $\bm a$ compared with the expected performance of the policy at state $s$. In the traffic scenario, agents collaborate to maximize the sum of future rewards $R_t = \sum \limits_{k=0}^{\infty} \gamma^{k} r_{t+k}$  by optimizing the joint policy $\boldsymbol \pi$.

\subsection{Observation}
\subsubsection{Observation Definition}
For heterogeneous methods, the observation of an agent $i$ corresponding to intersection $u$ is defined as the concatenation of the states of all incoming lanes at that intersection. Following prior study~\cite{gu2024large}, we adopt three types of features as the observation inputs: the number of waiting vehicles, the wave count representing the number of vehicles in the incoming lanes, and the current signal phase. Only incoming lanes are considered in the observation, since they provide direct information about approaching traffic flows that determine the signal control decisions. Formally, the $o^i$ is presented as:
 \begin{equation}
    o^{i}=\left\{\{\text {wait}[l], \text {wave}[l]\}_{l \in \text{InLanes}[u]}, \text {phase}[u]\right\},
\end{equation}
where $\text{InLanes}[u]$ denotes the set of incoming lanes at intersection $u$.

For homogeneous methods, the observation is designed to construct a unified feature representation across all intersections. To achieve this, we aggregate lane-level features according to the four directional approaches (North, South, West, East) at each intersection. For each approach, three functional lane types are considered, including left-turn, through, and right-turn. Each lane type contributes two numerical features: the wait count and wave count, resulting in 6 dimensions per approach and 24  dimensions in total. The current signal phase of the intersection is then appended, forming a fixed-length observation vector with 25 dimensions overall. For mixed lanes (e.g., left-through or through-right lanes), the corresponding lane features are evenly distributed across the related functional types. For instance, if a lane allows both left-turn and through movements, its wait and wave counts are divided by two and added equally to the left-turn and through dimensions. For T junctions, where certain approaches or lane types are absent, the corresponding feature entries are zero-padded to maintain consistent dimensionality across agents. Formally, the observation of agent $i$ at intersection $u$ is expressed as:
\begin{equation}
    o^{i}=\left\{\{\widehat{\text{wait}}[d, k], \widehat{\text{wave}}[d, k]\}_{d \in \mathcal{D}, k \in \mathcal{F}}, \text{phase}[u]\right\},
\end{equation}
where $\mathcal D=\{\text{N,E,S,W}\}$ denotes the set of directions, and $\mathcal F=\{\text{left-turn,through,right-turn}\}$ represents the functional lane types.

\subsubsection{Lane-level Missing Observation Modeling}
Lane-level measurements can be partially unavailable in practice, resulting in missing observations for some incoming lanes. To account for this factor, we introduce an observation missing rate $p_{\text{miss}}$, representing the probability that the features of a given incoming lane are unavailable. Each episode corresponds to one simulated hour of traffic signal control. At the beginning of each episode, for each lane $l \in \text{InLanes}[u]$, its corresponding wait and wave features are randomly masked with probability $p_{\text{miss}}$, and their values are set to zero for the entire episode. By resampling the missing lane set across episodes, this setting emulates diverse lane-level sensor availability patterns during training and enables robustness evaluation under varying missing configurations. The signal phase is assumed to be always observable and is therefore excluded from the masking process. Formally, for heterogeneous methods, the masked observation of agent $i$ corresponding to intersection $u$ is expressed as:

\begin{equation}
\begin{aligned}
\tilde{o}^{i}
&=
\left\{
\{\tilde{\text{wait}}[l],\ \tilde{\text{wave}}[l]\}_{l \in \text{InLanes}[u]},
\ \text{phase}[u]
\right\}, \\[4pt]
\tilde{\text{wait}}[l]
&=
\begin{cases}
0, & \text{with probability } p_{\text{miss}}, \\
\text{wait}[l], & \text{otherwise},
\end{cases} \\[4pt]
\tilde{\text{wave}}[l]
&=
\begin{cases}
0, & \text{with probability } p_{\text{miss}}, \\
\text{wave}[l], & \text{otherwise}.
\end{cases}
\end{aligned}
\end{equation}

Similarly, for homogeneous methods, masking is applied at the lane level before aggregating features into directional and functional categories. At the beginning of each episode, each incoming lane’s wait and wave entries are independently set to zero with probability $p_{\text{miss}}$. The remaining lane features are then aggregated into the $[d,k]$ dimensions obeying the same rule as in the observation definition. Thus, the masked observation for agent $i$ at intersection $u$ is
\begin{equation}
\tilde{o}^{i}
=
\left\{
\{\widehat{\tilde{\text{wait}}}[d,k],\ \widehat{\tilde{\text{wave}}}[d,k]\}_{d \in \mathcal{D},\, k \in \mathcal{F}},
\ \text{phase}[u]
\right\}.
\end{equation}

\begin{figure}[t]
    \centering
    \includegraphics[trim=300 150 300 160, clip,scale=0.6]{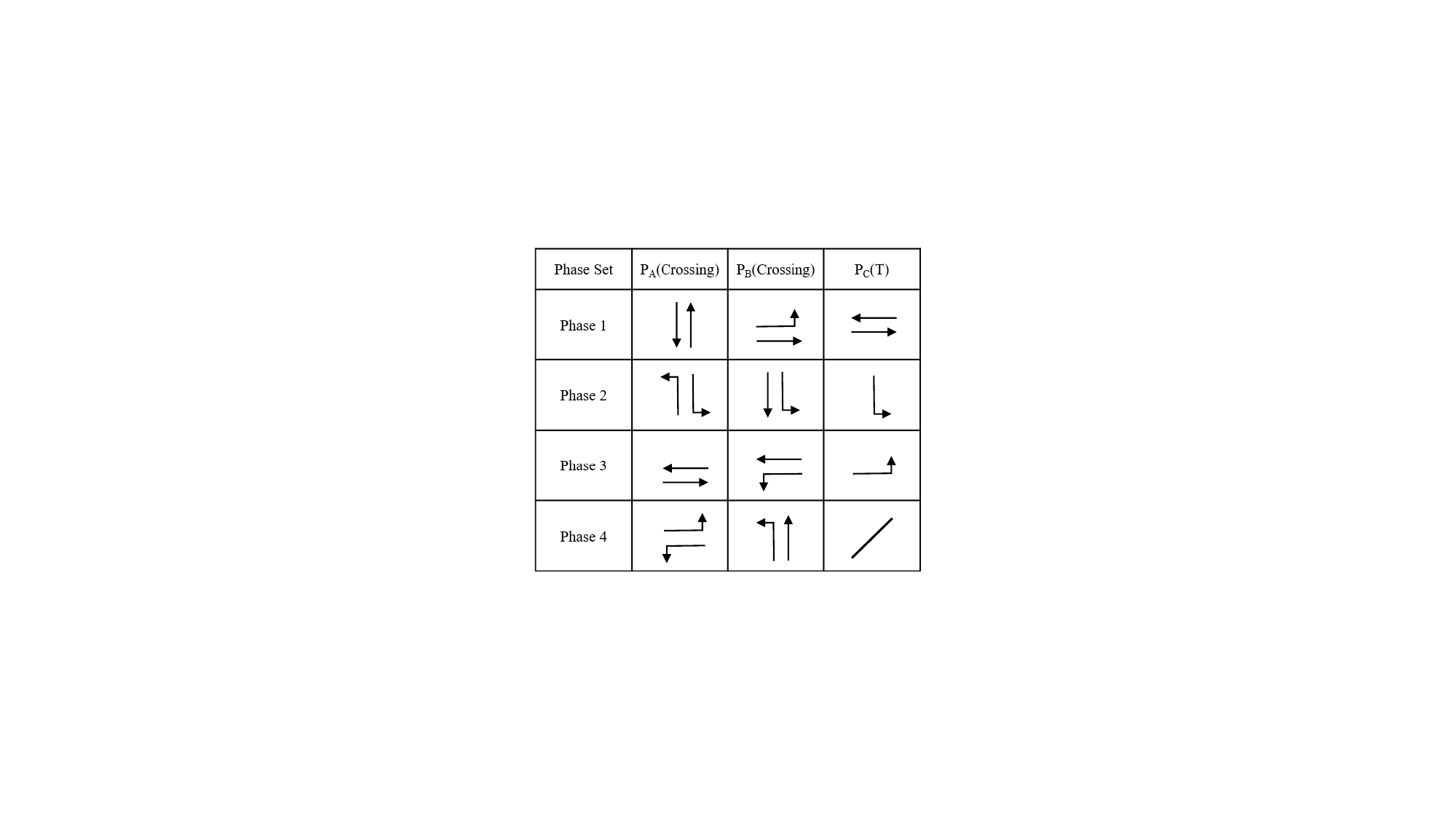}
    \caption{Figure of phase setting scheme for different intersection types: 
$\text{P}_\text{A}$ and $\text{P}_\text{B}$ correspond to two typical phase plans for crossing intersections, while $\text{P}_\text{C}$ represents a phase plan for a T intersection.}
    \label{phase_plans}
\end{figure}

\subsection{Action}
In real-world urban scenarios, intersections often exhibit different traffic volumes and geometric structures, which naturally lead to diverse phase setting strategies. To capture such heterogeneity, we introduce three phase setting schemes. Specifically, for crossing intersections, two alternative four-phase configurations are defined, denoted as $\text{P}_\text{A}$ and $\text{P}_\text{B}$; for T intersections, a three-phase configuration $\text{P}_\text{C}$ is adopted. It is assumed that right-turn movements are always allowed. The phase setting schemes of crossing intersections and T intersections $\text{P}_\text{A}$, $\text{P}_\text{B}$, and $\text{P}_\text{C}$ are illustrated in Fig.~\ref{phase_plans}. Formally, the action space of agent $i$ at intersection $u$ is defined as:
\begin{equation}
\mathcal A^i=
\begin{cases}
\text{P}_\text{A}\text{ or } \text{P}_\text{B},&\text{if }u\text{ is a crossing intersection,}\\
\text{P}_\text{C},&\text{if }u\text{ is a T intersection.}
\end{cases}
\end{equation}

\subsection{Reward}
In TSC problem, the design of the reward function plays a crucial role in guiding agents toward efficient traffic coordination. Various performance indicators, including waiting time, the number of waiting vehicles, delay, and travel speed, have been widely adopted in prior studies to characterize traffic efficiency. In this work, we employ the negative summation of the number of waiting vehicles on all incoming lanes as the reward. The instantaneous reward for agent $i$ at timestep $t$ is defined as:
\begin{equation}
    r_t^i=-\sum_{l \in \text{InLanes}[u]}\text{wait}_t[l],
\end{equation}
where $\text{wait}_t[l]$ denotes the number of waiting vehicles on the incoming lane $l$ of intersection  at time $t$.
The system-level reward at each timestep is obtained by aggregating the local rewards of all agents:
\begin{equation}
    r_t=\sum_{i=1}^{n}r_t^i.
\end{equation}
\begin{algorithm}[t]
\caption{The HALight Algorithm}
\label{alg:halight}
\SetAlgoLined 
\KwIn{Learning rate $\alpha$, batch size $B$, number of \\ 
agents $n$, episodes $K$, steps per episode $T$, traffic network $G_{het}$}

Initialize actor networks $\{\theta^i_0\}_{i=1}^n$, centralized critic $\phi_0$, and replay buffer $\mathcal{B}$\;
Determine the first-order neighbor sets $\mathcal{N}(i)$ for each agent $i$ from $G_{\text{het}}$\;

\For{$k = 0,1,\dots,K-1$}{
    Reset critic's and all actors' GRU hidden states\;
    Collect a trajectory using joint policy $\bm \pi_{\theta_k}$\;
    Push $\{(s_t,\tilde{o}_t^i, a_t^i, r_t, \tilde{o}_{t+1}^i,h_t^i,\text{info}_t^i),\forall t \in T\}$ in $\mathcal{B}$\;

    Sample $B$ trajectory segments from $\mathcal{B}$ \textbf{in temporal order} \tcp*{Required due to GRU}
    
    Compute advantages $\hat{A}(s,\bm{a})=\hat R-V_{\phi}(s)$ using the centralized critic with GAE\;
    
    Generate a random permutation $(i_1,i_2,\dots,i_n)$ of agent indices\;
    Set $M^{i_1}=1$\; 
    
    \For{$m = 1,2 \dots ,n$}{
        Compute importance sampling ratio $\omega^{i_m}$ according to Equation~\eqref{eq:importance_weight}\;
        Update actor $i_m$ to $\theta^{i_m}_{k+1}$ by maximizing:
        \[
        M^{i_{1:m}} \hat A(s,\bm{a})
        \cdot
        \min\!\big(
            \omega^{i_m},
            \operatorname{clip}(\omega^{i_m}, 1\pm\epsilon)
        \big).
        \]
        
        Update joint policy correction factor:
        \[
        M^{i_{1:m+1}}
        =
        \omega^{i_m} \, M^{i_{1:m}}.
        \]
    }
    
    Update critic parameters:
    \[
        \phi_{k+1} \gets 
        \arg\min_{\phi}
        \frac{1}{BT}
        \sum_{b=1}^{B}
        \sum_{t=0}^{T}
        \left(
            V_{\phi}(s_t) - \hat R_t
        \right)^2
    \]
}
\end{algorithm}
\section{HALight Architecture}
\label{sec:HALight}
\begin{figure*}[t]
    \centering
    \includegraphics[
        width=1.0\textwidth,
        clip
    ]{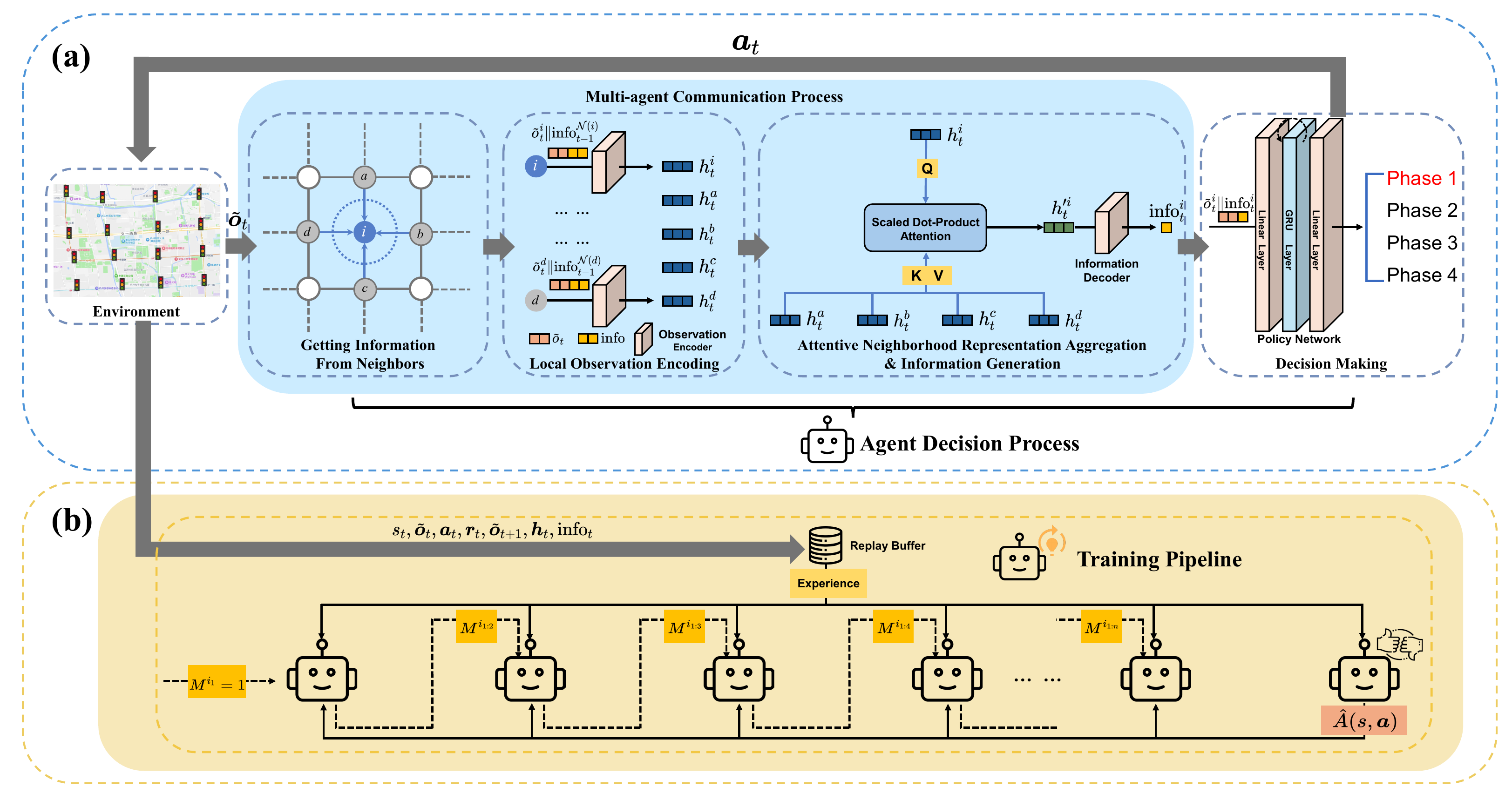}
    \caption{An overview of HALight, including (a) the agent decision process, which consists of local observation encoding, one-hop neighborhood communication, attentive aggregation, information generation, and decision making, and (b) the sequential training pipeline with replay buffer sampling and sequential updates.}
    \label{fig:halight_arch}
\end{figure*}

This section presents a heterogeneous MARL algorithm for TSC with lane-level missing observations, referred to as HALight. The overall framework of the proposed model is illustrated in Fig.~\ref{fig:halight_arch} and the pseudo-code of the HALight is presented in Algorithm~\ref{alg:halight}. The proposed HALight framework mainly consists of two key components: Communication-Enhanced Heterogeneous Actors and a Centralized Critic, which will be described in the following subsections.

\subsection{Communication-Enhanced Heterogeneous Actors}
Following the communication motivation of CCGM~\cite{peng2024enhancing}, HALight equips each heterogeneous actor with an explicit inter-agent communication module to enhance coordination under partial observability. Different from CCGM, we tailor the communication protocol to TSC by restricting message exchanges to one-hop physical neighbors and integrating the received messages via a lightweight dot-product attention mechanism, rather than relying on GAT-based aggregation and dynamic communication graph reconstruction. This design is motivated by the intrinsic locality of traffic interactions: adjacent intersections connected by road segments exhibit the strongest coupling in queue propagation and phase interference, and therefore provide the most informative spatial context for compensating missing observations while keeping the communication overhead scalable in large road networks.
\subsubsection{Local Observation Encoding}
At each timestep $t$, the encoder of agent $i$ processes its local observation together with the communication information received from its neighboring agents at the previous timestep. Let $\text{info}^j_{t-1}$ denote the message broadcast by neighbor $j$. The encoded latent representation is given by:
\begin{equation}
    h^i_t=f_{\text{enc}}^i(\operatorname{concat}(\tilde{o}_t^i,\{\text{info}^j_{t-1}\}_{j\in \mathcal N(i)})),
\end{equation}
where $f^i_{\text{enc}}(\cdot)$ denotes a single layer linear encoder, and $\mathcal N(i)$ represents the set of one-hop neighboring intersections in $G_{\text{het}}$. The generation process of $\text{info}^j_{t-1}$ will be elaborated later in this section.

\subsubsection{Attentive Neighborhood Representation Aggregation}
The latent representation $h_t^i$ reflects a preliminary understanding of the local environment for agent $i$ at timestep $t$. Considering that neighboring intersections exhibit distinct traffic conditions and exert varying degrees of influence on the target intersection, each agent is expected to selectively aggregate information from its neighbors rather than treating them equally. To achieve this, we adopt a dot-product attention mechanism with learnable linear projections. Specifically, the current hidden state $h_t^i$ is mapped to a query vector, while the neighbors' hidden states $\{h_t^j\}_{j \in \mathcal{N}(i)}$ are mapped to key and value vectors:
\begin{equation}
    q_t^i = W_q h_t^i,\qquad
k_t^j = W_k h_t^j,\qquad
v_t^j = W_v h_t^j,
\end{equation}
where $W_q, W_k, W_v \in \mathbb R^{d_{a}\times d_h}$ are training projection matrices and $d_{a}$ denotes the dimension of the attention vector. The normalized attention coefficient is then computed as
\begin{equation}
    \alpha_{ij} \;=\; 
\frac{\exp\!\big(q_t^i(k_t^j)^{\mathrm{T}}/ \sqrt{d_{a}}\big)}
{\sum_{j \in \mathcal{N}(i)} \exp\!\big(q_t^i(k_t^j)^{\mathrm{T}} / \sqrt{d_{a}}\big)}\,,
\end{equation}
$\alpha_{ij}$ measures the relative importance of neighbor $j$ to agent $i$. Thus, the enhanced neighborhood representation is obtained by the attention-weighted aggregation of values:
\begin{equation}
    h_{t}^{\prime i} = \sum_{j \in \mathcal{N}(i)} \alpha_{ij}\, v_t^j.
\end{equation}
Since $f^i_{\text{enc}}(\cdot)$ projects all agents’ inputs into a shared latent space of dimension $d_h$, the linear projections $(W_q, W_k, W_v)$ consistently map these representations into the common attention space $d_{a}$, enabling the dot-product operation to be applied in a principled manner.
\subsubsection{Information Generation}
After obtaining the enhanced neighborhood representation $h_{t}^{\prime i}$, each agent generates a piece of self-interested information through its communication decoder head, which is then transmitted to its neighboring agents and is simultaneously used for its own decision making. At this stage, the neighborhood information aggregated is condensed into a latent hypothesis $\text{info}_{t}^{i}$. This hypothesis reflects the action tendency of the sending agent, providing implicit cues that allow neighboring agents to make more informed inferences and coordinate their subsequent actions accordingly. Since the generated information encapsulates the agent’s behavioral tendency, its values are constrained within the range $[0,1]$, using a sigmoid activation function. Formally, the self-interested information of agent $i$ at timestep $t$ is defined as:
\begin{equation}
    \text{info}_{t}^{i}=\sigma\left(f^i_{\text{dec}}\left(h_{t}^{\prime i}\right)\right),
\end{equation}
where $f^i_{\text{dec}}(\cdot)$ denotes the information decoder that projects  $h_{t}^{\prime i}$ into a $d_{\text{info}}$ dimensional information vector and $\sigma(\cdot)$ represents the sigmoid activation function. This design ensures that the transmitted information remains bounded and interpretable, facilitating stable communication and mutual coordination among heterogeneous agents.

\subsubsection{Decision Making}
Having generated the self-interested communication information $\text{info}_t^i$, agent $i$ enters the action computation phase. A policy network is employed, consisting of a feedforward module, a gated recurrent unit (GRU)~\cite{dey2017gate}, and an action head. The GRU mitigates partial observability by maintaining a memory of past states, similar to recurrent reinforcement learning architectures.

At timestep $t$, the agent $i$ concatenates its local observation $\tilde{o}_t^i$ with the communication information $\text{info}_t^i$ to form a fused input vector, which is then processed by a two-layer feedforward module:
\begin{equation}
    x_t^i = \text{ReLU}\left(W_2 \, \text{ReLU}\left(W_1 \, \operatorname{concat}(\tilde{o}_t^i, \text{info}_t^i) + b_1\right) + b_2\right),
\end{equation}
where $W_{1,2}$ and $b_{1,2}$ are the parameters of the feedforward module. The output $x_t^i$ is then passed through the GRU to obtain the action feature $z_t^i$. Let $h_t^{\text{pol},i}$ denote the GRU hidden state of the policy network for agent $i$, the GRU processes the input as follows:
\begin{equation}
    z_t^i, h_t^{\text{pol},i} = \text{GRU}(x_t^i, h_{t-1}^{\text{pol},i}).
\end{equation}

The action feature $z_t^i$ is then passed to the action head to produce the policy distribution over the available phases:
\begin{equation}
\pi^i(a_t^i \mid \tilde{o}_t^i, \{\text{info}^j_{t-1}\}_{j\in \mathcal N(i)}) 
= \text{softmax}\big(W_{\text{act}} z_t^i + b_{\text{act}}\big),
\end{equation}
from which the action $a_t^i$ is sampled.
\subsection{Centralized Critic}
Following the CTDE paradigm, HALight employs a centralized critic to estimate value functions using joint information during training. The critic receives a global shared state that integrates not only each agent’s local observation but also the communication information produced by the actors. This design is motivated by the fact that the generated information vectors encode each agent’s action tendency and implicitly reflect the influence of neighborhood interactions. Concretely, at timestep $t$, each agent $i$ forms a compact local feature $g_t^i=\operatorname{concat}(\tilde{o}_t^i,\text{info}_t^i)$. The global state $s_t$ is obtained by concatenating all agents’ local features:
\begin{equation}
     s_t=\operatorname{concat}(g_t^1,g_t^2,\dots,g_t^{n}).
\end{equation}

The critic network processes $s_t$ through a two-layer feedforward module followed by a GRU, which helps improve the reliability of value estimation under partial observability. Formally, the critic update is given by:
\begin{equation}
    v_{t}, h_{t}^{c}=\text{GRU}\left(f_{\mathrm{MLP}}\left(s_{t}\right), h_{t-1}^{\text{c}}\right),
\end{equation}
where $f_{\text{MLP}(\cdot)}$ denotes the feedforward module, $v_t$ denotes the output feature vector produced by the GRU and $h_t^{\text{c}}$ is the critic’s recurrent hidden state.

Finally, the state value $V(s_t)$ is obtained through a linear output layer:
\begin{equation}
    V(s_t)=W_{v} v_{t}+b_{v},
\end{equation}
where $W_v$ and $b_v$ denote the parameters of the value head.
\subsection{Training Pipeline}

HALight adopts a HAPPO-style sequential policy update scheme for heterogeneous agents. 
Under full observability, the recurrent policy representation and communication modules can be regarded as components of individual policies. They enrich the policy inputs while preserving the factorized joint policy structure required by HAPPO. The convergence analysis of HALight under this setting is provided in Appendix~\ref{app:convergence}. 
Since the main task considers lane-level missing observations, the above theoretical guarantee is established only under full observability. The effectiveness of the sequential update and the proposed modules under partial observability is further validated through empirical evaluations.

The actor and critic update procedures are described below.

\begin{figure}[t]
    \centering
    \includegraphics[width=\linewidth]{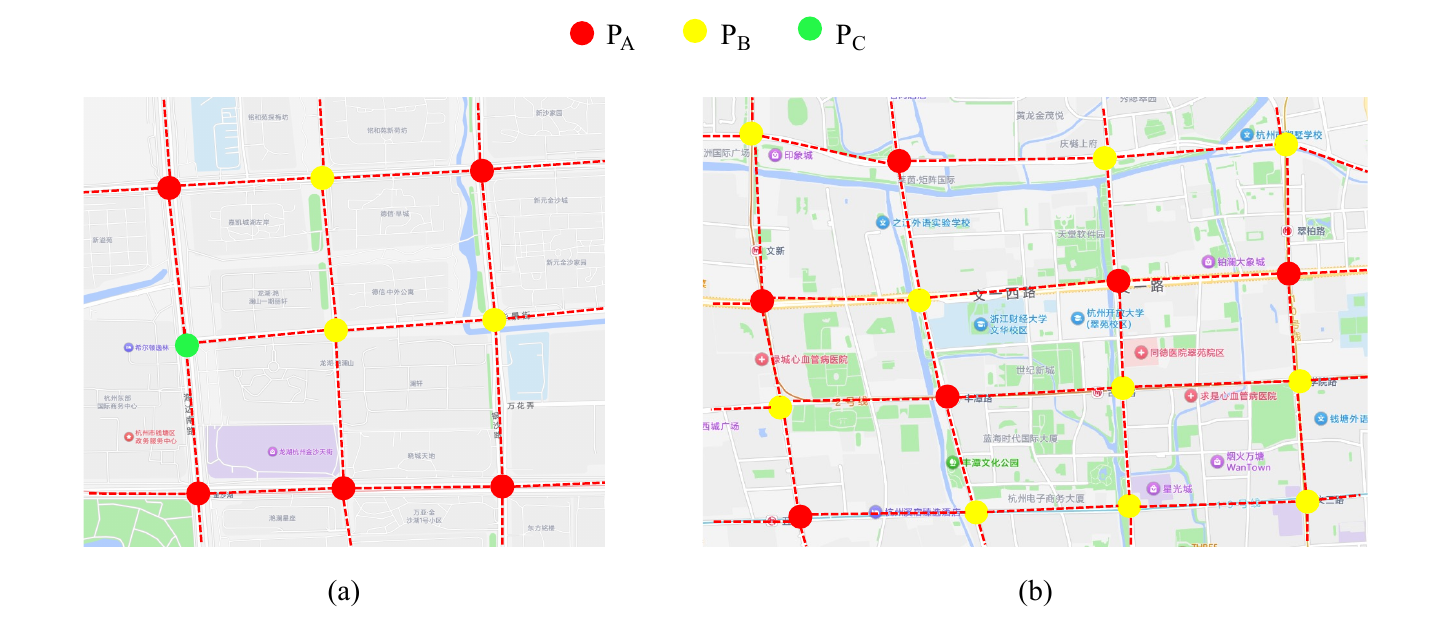} 
    \caption{Figures of the evaluated road networks, including (a) Xiasha road network and (b) Gudang road network.}
    \label{roadnet}
\end{figure}

\subsubsection{Actor Optimization}
At each training iteration, advantage estimates $\hat A(s,\bm a)$ for all agents are computed using generalized advantage estimation (GAE) and the centralized critic. Let agents be indexed as $i=1,2,\dots,n$, and denote their policy network by $\theta^1,\theta^2,\dots,\theta^n$. A random permutation $(i_1,i_2,\dots,i_n)$ of these agent indices is sampled, and the actors are updated sequentially following this order.

Consider the update of the $m$-th agent in the permutation, i.e., agent $i_m$. To ensure that the sequential updates are consistent with the improvement of the joint policy, HALight adopts the joint policy correction factor $M^{i_{1: m}}$, which accounts for the influence of previously updated agents on the current policy update:
\begin{equation}
    M^{i_{1: m}}=\prod_{q=1}^{m-1} \frac{\pi_{\theta_{k+1}^{i_q}}^{i_{q}}\left(a^{i_{q}} \mid \tilde{o}^{i_{q}},\left\{\text{info}^{j}\right\}_{j \in \mathcal{N}\left(i_{q}\right)}\right)}{\pi_{\theta_{k}^{i_q}}^{i_{q}}\left(a^{i_{q}} \mid \tilde{o}^{i_{q}},\left\{\text{info}^{j}\right\}_{j \in \mathcal{N}\left(i_{q}\right)}\right)},
\end{equation}
where $k$ denotes the policy optimization iteration, and $\theta_k^{i_q}$ and $\theta_{k+1}^{i_q}$ represent the policy parameters of agent $i_q$ before and after its sequential update, respectively. For $m=1$, the correction factor is defined as $M^{i_1}=1$. The policy of agent $i_m$ is then updated by maximizing the clipping objective scaled by this correction factor:
\begin{equation}
    \begin{split}
        &\theta  ^{i_m}_{k+1} \leftarrow \\
        &\arg \max _{\theta^{i_m}} 
        \mathbb{E}
        \Big[
        M^{i_{1:m}} \hat A(s,\bm{a})\,
        \min\big(
        \omega^{i_m},
        \operatorname{clip}(\omega^{i_m}, 1\pm\epsilon)
        \big)
        \Big],
    \end{split}
\end{equation}
where $\epsilon$ is the clipping parameter, the expectation is taken over 
$s \sim \rho_{\bm \pi_{\boldsymbol{\theta}_{k}}}$ and 
$\bm{a} \sim \bm \pi_{\boldsymbol{\theta}_{k}}$, 
and the importance sampling ratio $\omega^{i_m}$ is defined as:
\begin{equation}
    \omega^{i_m}
    =
    \frac{
    \pi_{\theta^{i_m}}^{i_m}
    \big(
    a^{i_m} \mid 
    \tilde{o}^{i_m},
    \{\text{info}^j\}_{j \in \mathcal{N}(i_m)}
    \big)
    }{
    \pi_{\theta^{i_m}_k}^{i_m}
    \big(
    a^{i_m} \mid 
    \tilde{o}^{i_m},
    \{\text{info}^j\}_{j \in \mathcal{N}(i_m)}
    \big)
    }.
\label{eq:importance_weight} 
\end{equation}
This completes the sequential update for the $m$-th actor, and the process continues until all agents in the permutation have been updated once within the training iteration.
\subsubsection{Critic Optimization}
The centralized critic is trained to approximate the state value function $V(s_t)$ based on the global shared state. During training, the critic parameters $\phi$ are optimized by minimizing the mean squared error between the predicted value and a bootstrapped return $\hat R_t$. The critic's loss function is given by:
\begin{equation}
    \mathcal{L}_{\text{critic}}(\phi)=\left( V_{\phi}(s_t) - \hat{R}_t \right)^2.
\end{equation}
The target return $\hat{R}_t$ is computed using GAE, and the critic update is performed after sequential actor updates.

\section{EXPERIMENTS AND RESULTS}
\label{sec:experiments}
In this section, we evaluate HALight on two real-world road networks, and compare the algorithm with representative homogeneous and heterogeneous baselines.
\subsection{Experiment Scenario}
We select two real-world road networks in Hangzhou for evaluation: a 3 $\times$ 3 grid road network in Xiasha and a 4 $\times$ 4 grid road network in Gudang. The Gudang network consists of 16 crossing intersections with varying structural configurations, while the Xiasha network contains 7 crossing intersections and 2 T intersections with different configurations. The road network topology and phase settings are illustrated in Fig.~\ref{roadnet}. The lane configurations of both networks are constructed according to real-world conditions, with intersection information obtained from AutoNavi\footnote{\url{https://ditu.amap.com}}. For each road network, we consider two traffic demand scenarios: moderate and high demand. The origin and destination of each vehicle are randomly sampled from the boundary edges of the network, and the travel path is determined by the shortest route between them. The road network and flow statistics are summarized in Table~\ref{tab:road_flow_stats}.

\begin{figure*}[t]
    \centering
    \includegraphics[
        width=\textwidth,
        clip
    ]{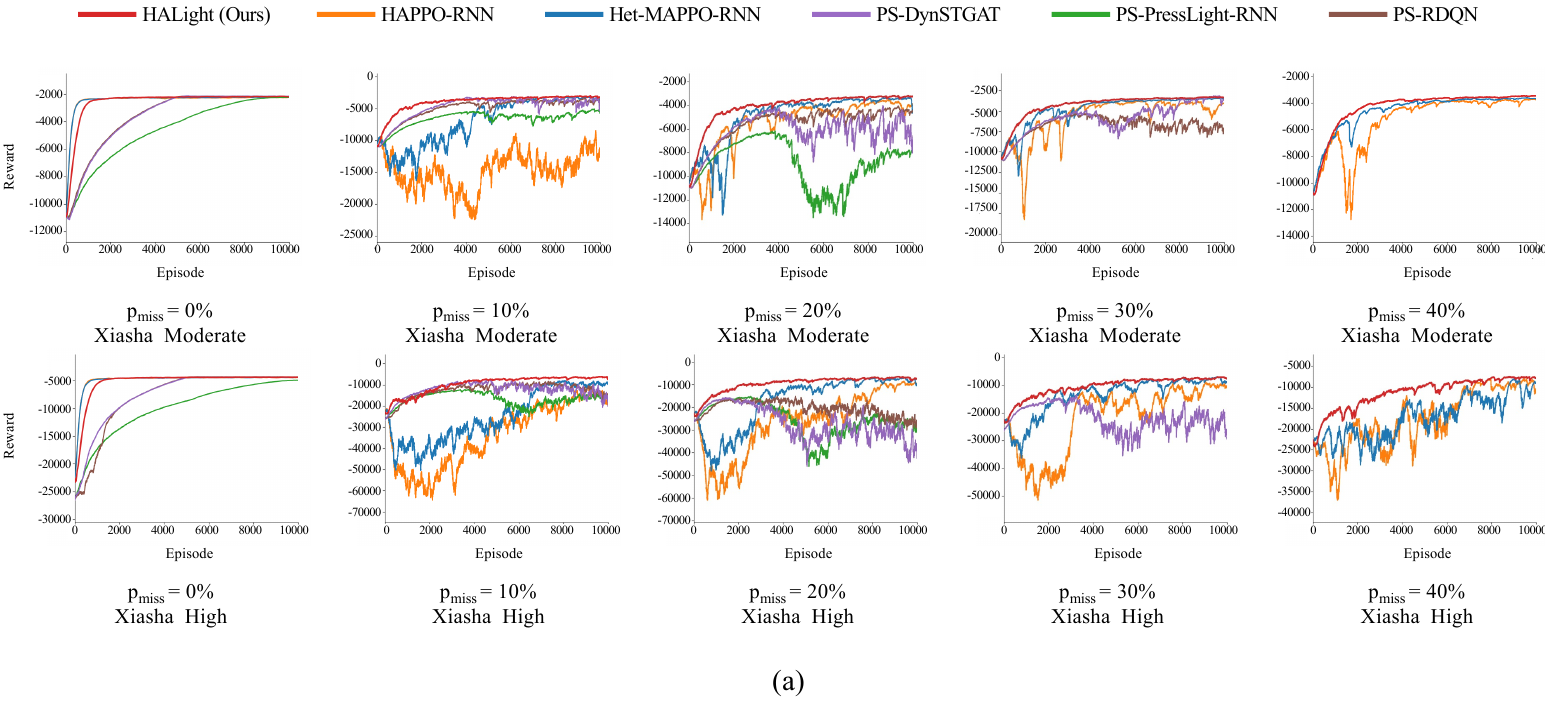}\\[0.3em]
    \includegraphics[
        width=\textwidth,
        clip
    ]{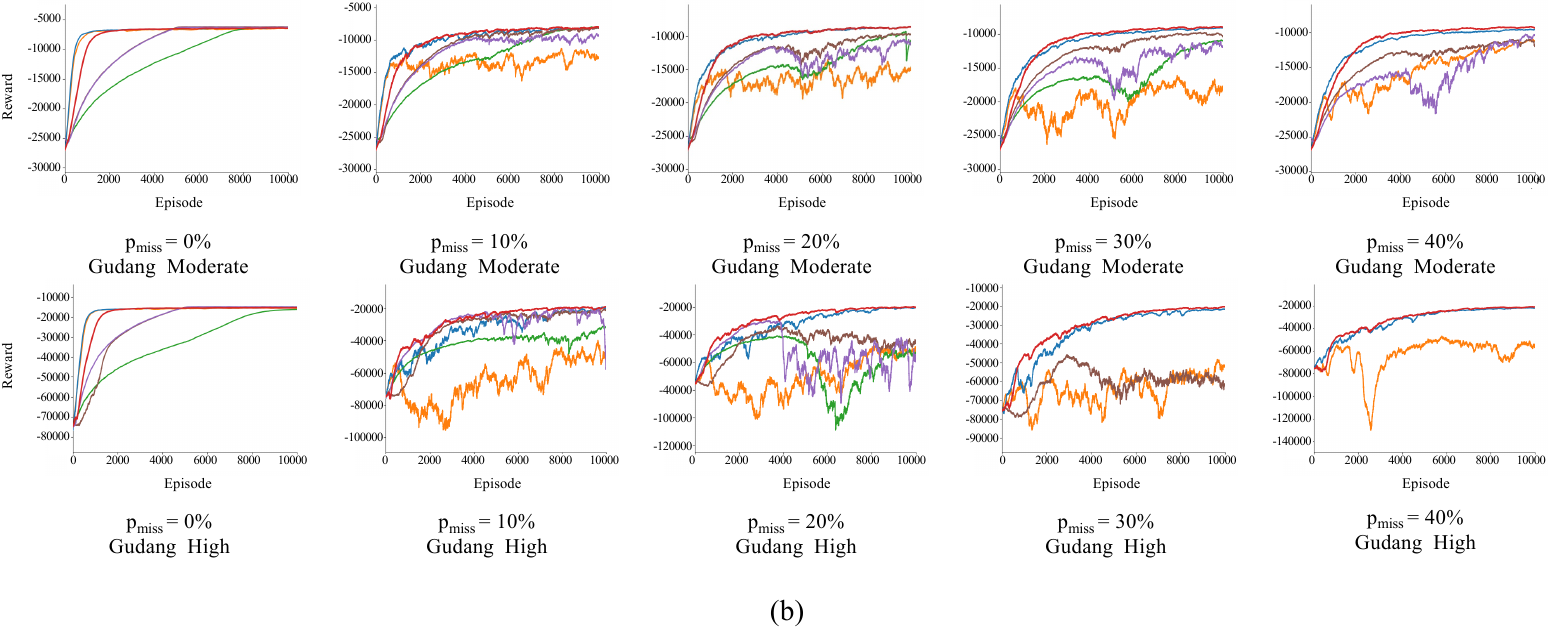}
    \caption{Learning curves of RL-based TSC methods on (a) the Xiasha road network and (b) the Gudang road network.}
    \label{fig:training_curves}
\end{figure*}

\begin{table}[h]
\caption{ROAD NETWORKS AND FLOW STATISTICS}
\label{tab:road_flow_stats}
\centering
\resizebox{\linewidth}{!}{
\begin{tabular}{|c|cc|l|}
\hline
\multirow{2}{*}{Road Networks} & \multicolumn{2}{l|}{Arrival Rate (vehicles/s)} & \multirow{2}{*}{Road Length (m)} \\ \cline{2-3}
          & \multicolumn{1}{c|}{Moderate} & High &                           \\ \hline
          Xiasha    & \multicolumn{1}{c|}{0.4}      & 0.6  & \multicolumn{1}{c|}{400} \\ \hline
          Gudang    & \multicolumn{1}{c|}{0.85}     & 1.5  & \multicolumn{1}{c|}{600} \\ \hline
\end{tabular}
}
\end{table}

\begin{table}[h]
\caption{HYPERPARAMETERS}
\label{tab:hyperparameters}
\centering
\resizebox{\linewidth}{!}{
\begin{tabular}{|c|c|c|}
\hline
Algorithms                  & Hyperparameters                            & Value                       \\ \hline
\multirow{6}{*}{DQN-based} & Learning rate $\alpha$                 & 0.0005                      \\ \cline{2-3} 
                            & Replay buffer size                     & 1,800,000                   \\ \cline{2-3} 
                            & Hidden size                            & 64                          \\ \cline{2-3} 
                            & $\epsilon$ decay steps                 & 1,800,000                   \\ \cline{2-3} 
                            & Batch size                             & 32                          \\ \cline{2-3} 
                            & Training sequence length                  & 16                          \\ \hline
\multirow{4}{*}{PPO-based} & Learning rate $\alpha_{\text{critic}}$ & 0.0005                      \\ \cline{2-3} 
                            & Learning rate $\alpha_{\text{actor}}$  & 0.00005                     \\ \cline{2-3} 
                            & Hidden size                            & 64 for actors, 128 for critic \\ \cline{2-3} 
                            & GAE $\lambda$                          & 0.9                         \\ \hline
\end{tabular}
}
\end{table}

\begin{table*}[t]
\centering
\caption{Overall performance of HALight and baseline methods under different $p_{\mathrm{miss}}$ on the Xiasha road network. The best results are highlighted in bold, and the second-best results are underlined. Results for methods that fail to converge are omitted.}
\label{tab:Xiasha results}
\scriptsize
\setlength{\tabcolsep}{2pt}
\resizebox{\textwidth}{!}{
\begin{tblr}{
  width = \textwidth,
  colspec = {Q[c,m,1.4cm] Q[c,m,0.6cm] *{10}{Q[c,m,1.1cm]}},
  cells = {c,m},
  row{1-3} = {font=\bfseries},
  cell{1}{1} = {r=3}{c,m},
  cell{1}{2} = {r=3}{c,m},
  cell{1}{3} = {c=10}{c,m},
  cell{2}{3} = {c=5}{c,m},
  cell{2}{8} = {c=5}{c,m},
  cell{4}{1}  = {r=4}{c,m},
  cell{8}{1}  = {r=4}{c,m},
  cell{12}{1} = {r=4}{c,m},
  cell{16}{1} = {r=4}{c,m},
  cell{20}{1} = {r=4}{c,m},
  cell{24}{1} = {r=4}{c,m},
  cell{28}{1} = {r=4}{c,m},
  hline{1,32} = {1.2pt},
  hline{4} = {0.8pt},
  hline{8,12,16,20,24,28} = {0.5pt},
  hline{2-3} = {3-12}{0.5pt},
  vline{3,8} = {1-31}{0.5pt},
}
Algorithms & Metric & Scenario & & & & & & & & & \\
           &        & Moderate Demand & & & & & High Demand & & & & \\
           &        & 0\% & 10\% & 20\% & 30\% & 40\% & 0\% & 10\% & 20\% & 30\% & 40\% \\

Fixed Time & ATT
& \ms{196.66}{2.11}
& \ms{196.66}{2.11}
& \underline{\ms{196.66}{2.11}}
& \ms{196.66}{2.11}
& \ms{196.66}{2.11}
& \ms{212.25}{6.12}
& \underline{\ms{212.25}{6.12}}
& \ms{212.25}{6.12}
& \underline{\ms{212.25}{6.12}}
& \ms{212.25}{6.12} \\

& AQL
& \ms{1.26}{0.04}
& \ms{1.26}{0.04}
& \underline{\ms{1.26}{0.04}}
& \ms{1.26}{0.04}
& \ms{1.26}{0.04}
& \ms{2.74}{0.31}
& \underline{\ms{2.74}{0.31}}
& \underline{\ms{2.74}{0.31}}
& \underline{\ms{2.74}{0.31}}
& \ms{2.74}{0.31} \\

& AWT
& \ms{28.83}{1.05}
& \ms{28.83}{1.05}
& \ms{28.83}{1.05}
& \ms{28.83}{1.05}
& \ms{28.83}{1.05}
& \ms{43.50}{5.31}
& \ms{43.50}{5.31}
& \ms{43.50}{5.31}
& \ms{43.50}{5.31}
& \ms{43.50}{5.31} \\

& TP
& \ms{1358.49}{4.11}
& \underline{\ms{1358.49}{4.11}}
& \underline{\ms{1358.49}{4.11}}
& \ms{1358.49}{4.11}
& \ms{1358.49}{4.11}
& \ms{2025.02}{8.71}
& \underline{\ms{2025.02}{8.71}}
& \underline{\ms{2025.02}{8.71}}
& \underline{\ms{2025.02}{8.71}}
& \textbf{\ms{2025.02}{8.71}} \\

PS-RDQN & ATT
& \ms{183.63}{1.53}
& \underline{\ms{195.28}{12.10}}
& \ms{206.57}{28.24}
& \ms{211.78}{17.13}
& -
& \ms{189.92}{1.56}
& \ms{228.30}{94.16}
& \ms{288.31}{163.66}
& -
& - \\

& AQL
& \ms{0.71}{0.02}
& \underline{\ms{1.22}{0.54}}
& \ms{1.73}{1.31}
& \ms{1.97}{0.81}
& -
& \ms{1.38}{0.03}
& \ms{3.98}{6.42}
& \ms{8.05}{11.15}
& -
& - \\

& AWT
& \ms{14.46}{0.28}
& \ms{26.30}{8.66}
& \ms{36.91}{19.05}
& \ms{44.87}{14.46}
& -
& \ms{20.01}{0.68}
& \ms{42.62}{23.53}
& \ms{65.92}{30.91}
& -
& - \\

& TP
& \ms{1363.50}{2.87}
& \ms{1357.71}{10.74}
& \ms{1349.36}{34.74}
& \ms{1348.17}{25.72}
& -
& \ms{2040.68}{3.91}
& \ms{1988.93}{168.02}
& \ms{1895.70}{296.35}
& -
& - \\

\makecell{PS-PressLight\\-RNN} & ATT
& \underline{\ms{183.36}{1.79}}
& \ms{206.81}{20.34}
& \ms{223.32}{9.88}
& -
& -
& \ms{191.64}{1.77}
& \ms{231.91}{43.80}
& \ms{276.12}{76.34}
& -
& - \\

& AQL
& \ms{0.70}{0.02}
& \ms{1.74}{0.90}
& \ms{2.47}{0.45}
& -
& -
& \ms{1.51}{0.05}
& \ms{4.14}{2.90}
& \ms{7.02}{5.10}
& -
& - \\

& AWT
& \ms{14.38}{0.44}
& \ms{40.76}{20.11}
& \ms{59.93}{10.11}
& -
& -
& \ms{22.09}{0.74}
& \ms{62.85}{28.25}
& \ms{95.67}{25.69}
& -
& - \\

& TP
& \ms{1363.56}{3.09}
& \ms{1353.27}{15.73}
& \ms{1345.18}{8.54}
& -
& -
& \ms{2039.88}{4.67}
& \ms{2005.60}{66.58}
& \ms{1955.37}{140.74}
& -
& - \\

PS-DynSTGAT & ATT
& \ms{183.47}{1.20}
& \ms{212.84}{64.78}
& \ms{242.74}{154.20}
& \ms{219.61}{49.05}
& -
& \ms{189.75}{1.72}
& \ms{229.79}{117.60}
& \ms{371.22}{272.65}
& \ms{259.14}{141.07}
& - \\

& AQL
& \ms{0.70}{0.02}
& \ms{2.05}{3.02}
& \ms{3.42}{7.12}
& \ms{2.32}{2.23}
& -
& \underline{\ms{1.37}{0.04}}
& \ms{4.10}{8.01}
& \ms{13.70}{18.39}
& \ms{5.99}{9.37}
& - \\

& AWT
& \underline{\ms{14.37}{0.29}}
& \ms{37.06}{29.80}
& \ms{33.79}{28.80}
& \ms{44.41}{28.89}
& -
& \ms{19.58}{0.48}
& \ms{37.39}{40.96}
& \ms{62.74}{34.04}
& \ms{47.12}{37.34}
& - \\

& TP
& \ms{1363.97}{3.46}
& \ms{1331.45}{83.07}
& \ms{1301.15}{172.08}
& \ms{1330.37}{62.65}
& -
& \ms{2040.83}{4.55}
& \ms{1974.78}{213.61}
& \ms{1744.62}{467.12}
& \ms{1927.89}{253.89}
& - \\

\makecell{Het-MAPPO\\-RNN} & ATT
& \ms{183.55}{1.55}
& \ms{197.48}{42.31}
& \ms{197.16}{23.02}
& \underline{\ms{191.80}{2.74}}
& \underline{\ms{193.14}{2.75}}
& \underline{\ms{189.49}{1.71}}
& \ms{215.31}{45.11}
& \underline{\ms{210.94}{45.52}}
& \ms{212.28}{56.58}
& \underline{\ms{209.75}{20.22}} \\

& AQL
& \underline{\ms{0.69}{0.02}}
& \ms{1.33}{1.95}
& \ms{1.31}{1.04}
& \underline{\ms{1.06}{0.09}}
& \underline{\ms{1.12}{0.11}}
& \underline{\ms{1.37}{0.04}}
& \ms{3.10}{3.02}
& \ms{2.78}{3.04}
& \ms{2.85}{3.88}
& \underline{\ms{2.67}{1.37}} \\

& AWT
& \ms{14.40}{0.35}
& \underline{\ms{23.20}{11.03}}
& \underline{\ms{26.23}{18.28}}
& \underline{\ms{23.41}{2.55}}
& \underline{\ms{24.63}{2.38}}
& \underline{\ms{19.48}{0.67}}
& \underline{\ms{34.12}{22.04}}
& \underline{\ms{35.36}{22.37}}
& \underline{\ms{37.99}{18.24}}
& \ms{39.92}{14.51} \\

& TP
& \underline{\ms{1364.00}{3.93}}
& \ms{1352.11}{49.95}
& \ms{1354.84}{22.43}
& \underline{\ms{1360.64}{3.99}}
& \underline{\ms{1360.03}{4.63}}
& \underline{\ms{2041.04}{4.20}}
& \ms{2009.59}{72.74}
& \ms{2013.30}{84.32}
& \ms{2017.34}{106.53}
& \ms{2022.33}{32.82} \\

HAPPO-RNN & ATT
& \ms{183.66}{1.50}
& \ms{249.52}{130.45}
& \underline{\ms{196.66}{12.01}}
& \ms{195.33}{7.27}
& \ms{193.73}{2.23}
& \ms{189.70}{1.80}
& \ms{261.38}{124.99}
& \ms{217.34}{49.87}
& \ms{220.64}{70.00}
& \ms{213.30}{55.15} \\

& AQL
& \underline{\ms{0.69}{0.02}}
& \ms{3.67}{5.91}
& \ms{1.28}{0.55}
& \ms{1.21}{0.31}
& \ms{1.14}{0.07}
& \underline{\ms{1.37}{0.05}}
& \ms{6.15}{8.32}
& \ms{3.19}{3.27}
& \ms{3.41}{4.63}
& \ms{2.91}{3.77} \\

& AWT
& \ms{14.41}{0.32}
& \ms{38.79}{20.95}
& \ms{26.74}{9.48}
& \ms{27.05}{6.76}
& \ms{25.38}{1.95}
& \ms{19.56}{0.74}
& \ms{51.89}{36.41}
& \ms{41.58}{21.11}
& \ms{41.40}{17.91}
& \underline{\ms{38.14}{15.55}} \\

& TP
& \ms{1363.98}{3.79}
& \ms{1299.85}{147.03}
& \ms{1356.70}{10.35}
& \ms{1358.48}{6.89}
& \ms{1359.35}{4.30}
& \ms{2040.72}{4.19}
& \ms{1936.53}{205.48}
& \ms{2010.26}{81.58}
& \ms{2003.56}{123.91}
& \ms{2015.76}{101.30} \\

HALight & ATT
& \textbf{\ms{183.00}{1.78}}
& \textbf{\ms{189.88}{6.73}}
& \textbf{\ms{189.87}{2.79}}
& \textbf{\ms{190.74}{2.43}}
& \textbf{\ms{191.98}{2.32}}
& \textbf{\ms{189.39}{1.78}}
& \textbf{\ms{199.71}{11.82}}
& \textbf{\ms{199.25}{8.70}}
& \textbf{\ms{208.02}{24.15}}
& \textbf{\ms{208.02}{28.15}} \\

& AQL
& \textbf{\ms{0.68}{0.02}}
& \textbf{\ms{0.98}{0.30}}
& \textbf{\ms{0.98}{0.09}}
& \textbf{\ms{1.01}{0.07}}
& \textbf{\ms{1.07}{0.07}}
& \textbf{\ms{1.36}{0.05}}
& \textbf{\ms{2.04}{0.78}}
& \textbf{\ms{1.99}{0.59}}
& \textbf{\ms{2.57}{1.62}}
& \textbf{\ms{2.57}{1.95}} \\

& AWT
& \textbf{\ms{14.33}{0.48}}
& \textbf{\ms{20.80}{4.54}}
& \textbf{\ms{21.42}{2.57}}
& \textbf{\ms{22.20}{1.95}}
& \textbf{\ms{23.61}{1.93}}
& \textbf{\ms{19.43}{0.73}}
& \textbf{\ms{28.64}{9.22}}
& \textbf{\ms{29.38}{6.83}}
& \textbf{\ms{37.77}{15.88}}
& \textbf{\ms{36.63}{9.99}} \\

& TP
& \textbf{\ms{1364.01}{3.67}}
& \textbf{\ms{1360.44}{8.24}}
& \textbf{\ms{1361.41}{4.00}}
& \textbf{\ms{1360.99}{3.94}}
& \textbf{\ms{1360.32}{4.30}}
& \textbf{\ms{2041.17}{4.74}}
& \textbf{\ms{2033.02}{13.65}}
& \textbf{\ms{2033.26}{21.35}}
& \textbf{\ms{2025.08}{38.37}}
& \underline{\ms{2023.20}{64.04}} \\

\end{tblr}
}
\end{table*}

\begin{table*}[t]
\centering
\caption{Overall performance of HALight and baseline methods under different $p_{\mathrm{miss}}$ on the Gudang road network. The best results are highlighted in bold, and the second-best results are underlined. Results for methods that fail to converge are omitted.}
\label{tab:Gudang results}
\scriptsize
\setlength{\tabcolsep}{2pt}
\resizebox{\textwidth}{!}{
\begin{tblr}{
  width = \textwidth,
  colspec = {Q[c,m,1.4cm] Q[c,m,0.6cm] *{10}{Q[c,m,1.1cm]}},
  cells = {c,m},
  row{1-3} = {font=\bfseries},
  cell{1}{1} = {r=3}{c,m},
  cell{1}{2} = {r=3}{c,m},
  cell{1}{3} = {c=10}{c,m},
  cell{2}{3} = {c=5}{c,m},
  cell{2}{8} = {c=5}{c,m},
  cell{4}{1}  = {r=4}{c,m},
  cell{8}{1}  = {r=4}{c,m},
  cell{12}{1} = {r=4}{c,m},
  cell{16}{1} = {r=4}{c,m},
  cell{20}{1} = {r=4}{c,m},
  cell{24}{1} = {r=4}{c,m},
  cell{28}{1} = {r=4}{c,m},
  hline{1,32} = {1.2pt},
  hline{4} = {0.8pt},
  hline{8,12,16,20,24,28} = {0.5pt},
  hline{2-3} = {3-12}{0.5pt},
  vline{3,8} = {1-31}{0.5pt},
}
Algorithms & Metric & Scenario & & & & & & & & & \\
           &        & Moderate Demand & & & & & High Demand & & & & \\
           &        & 0\% & 10\% & 20\% & 30\% & 40\% & 0\% & 10\% & 20\% & 30\% & 40\% \\

Fixed Time & ATT
& \ms{308.19}{1.73}
& \ms{308.19}{1.73}
& \ms{308.19}{1.73}
& \ms{308.19}{1.73}
& \ms{308.19}{1.73}
& \ms{332.98}{5.29}
& \ms{332.98}{5.29}
& \ms{332.98}{5.29}
& \ms{332.98}{5.29}
& \ms{332.98}{5.29} \\

& AQL
& \ms{1.80}{0.03}
& \ms{1.80}{0.03}
& \ms{1.80}{0.03}
& \ms{1.80}{0.03}
& \ms{1.80}{0.03}
& \ms{5.11}{0.39}
& \ms{5.11}{0.39}
& \ms{5.11}{0.39}
& \ms{5.11}{0.39}
& \ms{5.11}{0.39} \\

& AWT
& \ms{33.24}{0.59}
& \ms{33.24}{0.59}
& \ms{33.24}{0.59}
& \ms{33.24}{0.59}
& \ms{33.24}{0.59}
& \ms{55.55}{4.29}
& \ms{55.55}{4.29}
& \ms{55.55}{4.29}
& \ms{55.55}{4.29}
& \ms{55.55}{4.29} \\

& TP
& \ms{2784.86}{6.83}
& \ms{2784.86}{6.83}
& \ms{2784.86}{6.83}
& \ms{2784.86}{6.83}
& \ms{2784.86}{6.83}
& \ms{4856.63}{19.56}
& \ms{4856.63}{19.56}
& \ms{4856.63}{19.56}
& \ms{4856.63}{19.56}
& \ms{4856.63}{19.56} \\

PS-RDQN & ATT
& \ms{295.26}{1.50}
& \ms{301.11}{4.64}
& \ms{305.64}{4.74}
& \ms{306.61}{3.38}
& \ms{311.98}{5.23}
& \ms{304.21}{1.74}
& \ms{317.45}{29.99}
& \ms{359.60}{62.35}
& \ms{383.98}{49.86}
& - \\

& AQL
& \underline{\ms{1.13}{0.02}}
& \ms{1.46}{0.23}
& \ms{1.70}{0.24}
& \ms{1.75}{0.17}
& \ms{2.04}{0.38}
& \ms{2.67}{0.05}
& \ms{3.94}{3.22}
& \ms{8.14}{6.66}
& \ms{10.57}{5.45}
& - \\

& AWT
& \ms{19.13}{0.30}
& \ms{25.73}{4.63}
& \ms{30.57}{4.61}
& \ms{32.12}{3.71}
& \ms{37.39}{5.70}
& \ms{26.54}{0.48}
& \ms{36.85}{7.39}
& \ms{64.22}{15.56}
& \ms{85.86}{17.75}
& - \\

& TP
& \ms{2797.18}{5.14}
& \ms{2790.68}{8.78}
& \ms{2787.20}{9.90}
& \ms{2785.74}{7.00}
& \ms{2780.35}{9.90}
& \ms{4919.06}{7.61}
& \ms{4880.24}{148.62}
& \ms{4726.93}{327.41}
& \ms{4648.19}{290.50}
& - \\

\makecell{PS-PressLight\\-RNN} & ATT
& \ms{295.46}{1.08}
& \ms{301.37}{2.99}
& \ms{312.96}{25.93}
& \ms{308.18}{2.75}
& -
& \ms{305.63}{1.64}
& \ms{332.50}{9.66}
& \ms{363.46}{21.51}
& -
& - \\

& AQL
& \ms{1.14}{0.02}
& \ms{1.47}{0.15}
& \ms{2.11}{1.56}
& \ms{1.83}{0.13}
& -
& \ms{2.80}{0.04}
& \ms{5.26}{0.90}
& \ms{8.20}{2.16}
& -
& - \\

& AWT
& \ms{19.31}{0.32}
& \ms{25.79}{2.95}
& \ms{34.55}{3.64}
& \ms{33.36}{2.53}
& -
& \ms{28.43}{0.45}
& \ms{57.34}{8.71}
& \ms{85.52}{14.03}
& -
& - \\

& TP
& \ms{2796.74}{6.68}
& \ms{2791.01}{7.88}
& \ms{2773.88}{68.92}
& \ms{2784.74}{8.63}
& -
& \ms{4918.19}{8.99}
& \ms{4866.56}{27.39}
& \ms{4786.05}{86.31}
& -
& - \\

PS-DynSTGAT & ATT
& \textbf{\ms{294.86}{1.54}}
& \ms{303.49}{7.01}
& \ms{315.73}{23.49}
& \ms{309.05}{8.98}
& \ms{307.61}{4.38}
& \textbf{\ms{303.86}{1.64}}
& \ms{351.54}{90.62}
& \ms{365.03}{70.41}
& -
& - \\

& AQL
& \textbf{\ms{1.09}{0.02}}
& \ms{1.60}{0.40}
& \ms{2.27}{1.35}
& \ms{1.89}{0.52}
& \ms{1.80}{0.23}
& \textbf{\ms{2.62}{0.04}}
& \ms{7.50}{9.41}
& \ms{8.72}{7.28}
& -
& - \\

& AWT
& \underline{\ms{19.08}{0.33}}
& \ms{27.82}{6.03}
& \ms{36.63}{12.14}
& \ms{34.05}{6.80}
& \ms{32.84}{4.08}
& \underline{\ms{26.52}{0.43}}
& \ms{38.83}{14.09}
& \ms{59.58}{21.36}
& -
& - \\

& TP
& \underline{\ms{2797.32}{4.78}}
& \ms{2786.57}{14.20}
& \ms{2767.68}{56.49}
& \ms{2782.67}{17.83}
& \ms{2784.43}{9.30}
& \underline{\ms{4919.42}{8.02}}
& \ms{4700.91}{449.37}
& \ms{4678.04}{349.59}
& -
& - \\

\makecell{Het-MAPPO\\-RNN} & ATT
& \ms{295.55}{1.32}
& \underline{\ms{301.01}{2.90}}
& \underline{\ms{302.90}{5.18}}
& \underline{\ms{303.15}{2.01}}
& \underline{\ms{305.04}{2.52}}
& \ms{304.76}{1.37}
& \underline{\ms{312.88}{11.35}}
& \underline{\ms{313.93}{8.85}}
& \underline{\ms{316.23}{7.97}}
& \underline{\ms{316.96}{5.42}} \\

& AQL
& \ms{1.14}{0.02}
& \underline{\ms{1.45}{0.13}}
& \underline{\ms{1.55}{0.27}}
& \underline{\ms{1.56}{0.08}}
& \underline{\ms{1.66}{0.11}}
& \ms{2.66}{0.03}
& \underline{\ms{3.45}{1.15}}
& \underline{\ms{3.55}{0.87}}
& \underline{\ms{3.78}{0.78}}
& \underline{\ms{3.84}{0.50}} \\

& AWT
& \ms{19.14}{0.13}
& \underline{\ms{25.55}{2.78}}
& \underline{\ms{27.31}{3.91}}
& \underline{\ms{27.94}{1.72}}
& \underline{\ms{30.17}{2.36}}
& \ms{26.58}{0.40}
& \underline{\ms{34.69}{6.12}}
& \underline{\ms{36.23}{5.09}}
& \underline{\ms{38.58}{5.04}}
& \underline{\ms{39.94}{4.54}} \\

& TP
& \ms{2797.29}{8.79}
& \underline{\ms{2791.14}{7.33}}
& \underline{\ms{2790.06}{10.31}}
& \underline{\ms{2790.18}{6.94}}
& \underline{\ms{2787.90}{7.60}}
& \ms{4919.23}{9.42}
& \underline{\ms{4898.08}{54.14}}
& \underline{\ms{4900.63}{28.24}}
& \underline{\ms{4893.63}{35.90}}
& \underline{\ms{4895.01}{20.09}} \\

HAPPO-RNN & ATT
& \ms{295.48}{1.48}
& \ms{313.09}{13.38}
& \ms{320.79}{17.08}
& \ms{329.64}{27.38}
& \ms{311.32}{4.70}
& \ms{304.95}{1.72}
& \ms{363.17}{56.54}
& \ms{367.26}{48.88}
& \ms{372.54}{47.81}
& \ms{391.52}{43.06} \\

& AQL
& \ms{1.15}{0.02}
& \ms{2.10}{0.74}
& \ms{2.53}{0.96}
& \ms{3.02}{1.55}
& \ms{2.01}{0.25}
& \ms{2.72}{0.05}
& \ms{8.49}{5.78}
& \ms{8.86}{5.08}
& \ms{9.33}{4.97}
& \ms{11.12}{4.39} \\

& AWT
& \ms{19.26}{0.32}
& \ms{37.43}{10.91}
& \ms{43.31}{9.90}
& \ms{50.46}{13.48}
& \ms{36.32}{4.17}
& \ms{27.42}{0.49}
& \ms{63.20}{23.51}
& \ms{68.73}{17.44}
& \ms{73.92}{20.87}
& \ms{94.93}{23.64} \\

& TP
& \ms{2797.17}{6.90}
& \ms{2775.13}{25.65}
& \ms{2764.20}{34.77}
& \ms{2754.04}{49.75}
& \ms{2780.94}{9.16}
& \ms{4918.90}{8.96}
& \ms{4676.15}{289.85}
& \ms{4684.21}{256.17}
& \ms{4693.80}{232.52}
& \ms{4650.36}{197.91} \\

HALight & ATT
& \underline{\ms{295.20}{1.55}}
& \textbf{\ms{299.93}{2.50}}
& \textbf{\ms{301.77}{2.46}}
& \textbf{\ms{302.87}{2.11}}
& \textbf{\ms{304.56}{2.39}}
& \underline{\ms{304.17}{1.75}}
& \textbf{\ms{310.37}{4.40}}
& \textbf{\ms{312.87}{4.00}}
& \textbf{\ms{313.06}{3.20}}
& \textbf{\ms{314.33}{3.32}} \\

& AQL
& \underline{\ms{1.13}{0.02}}
& \textbf{\ms{1.39}{0.12}}
& \textbf{\ms{1.48}{0.11}}
& \textbf{\ms{1.54}{0.08}}
& \textbf{\ms{1.64}{0.09}}
& \underline{\ms{2.65}{0.05}}
& \textbf{\ms{3.23}{0.39}}
& \textbf{\ms{3.45}{0.33}}
& \textbf{\ms{3.48}{0.25}}
& \textbf{\ms{3.58}{0.26}} \\

& AWT
& \textbf{\ms{19.06}{0.33}}
& \textbf{\ms{24.18}{2.03}}
& \textbf{\ms{26.41}{2.27}}
& \textbf{\ms{27.46}{1.44}}
& \textbf{\ms{29.79}{1.89}}
& \textbf{\ms{26.48}{0.51}}
& \textbf{\ms{32.70}{2.99}}
& \textbf{\ms{35.77}{3.33}}
& \textbf{\ms{36.14}{2.56}}
& \textbf{\ms{37.40}{2.26}} \\

& TP
& \textbf{\ms{2797.35}{6.77}}
& \textbf{\ms{2791.99}{7.46}}
& \textbf{\ms{2791.51}{6.60}}
& \textbf{\ms{2790.27}{7.53}}
& \textbf{\ms{2788.10}{6.98}}
& \textbf{\ms{4919.98}{8.97}}
& \textbf{\ms{4907.28}{16.11}}
& \textbf{\ms{4903.32}{13.26}}
& \textbf{\ms{4904.36}{11.91}}
& \textbf{\ms{4901.45}{13.56}} \\

\end{tblr}
}
\end{table*}

\subsection{Experimental Setup}
\subsubsection{Comparison Methods}
For comparison, we select the following baseline methods, categorized into homogeneous and heterogeneous methods, as well as a non-RL method. To mitigate the effects of partial observability, all baseline methods incorporate a recurrent neural network  within the policy network. To adapt to heterogeneous road networks, the homogeneous methods utilize observation combination, action masking and parameter sharing. In particular, action masking sets the Q-values of illegal actions to $-\infty$ to prevent them from being selected.
\begin{itemize}
    \item \textbf{Fixed Time}: A rule-based, non-RL method that operates with fixed time signals for each intersection, regardless of traffic conditions. This method serves as a baseline to compare the performance of RL-based methods against traditional TSC strategies.        
    
    \item \textbf{PS-RDQN}~\cite{hausknecht2015deep}: A homogeneous approach based on Q-learning, where all agents share a common deep Q-network and are trained centrally while executing independently. To alleviate partial observability, RDQN integrates a recurrent neural network into the Q-network.

    \item \textbf{PS-PressLight-RNN}~\cite{wei2019presslight}: We adopt a parameter sharing, CTDE version of PressLight as a homogeneous method for comparison. PressLight optimizes agents by using the maximum pressure strategy, where the pressure at each intersection is defined as the difference between the incoming and outgoing traffic flow capacities.

    \item \textbf{PS-DynSTGAT}~\cite{wu2021dynstgat}: A parameter sharing homogeneous spatio-temporal graph attention baseline for TSC, which captures spatial dependencies among intersections through graph attention and temporal dependencies from historical traffic states. Action masks are used to handle different feasible phase sets under the shared action space.



    \item \textbf{Het-MAPPO-RNN}~\cite{yu2022surprising}: A heterogeneous variant of MAPPO, where each intersection is equipped with an independently modeled actor network to accommodate intersection-specific observation and action spaces, while all agents share a centralized critic for value estimation.
    
    \item \textbf{HAPPO-RNN}~\cite{kuba2022trust}: A PPO based method for heterogeneous agents. It uses the multi-agent advantage decomposition lemma and sequential updates to support monotonic joint policy improvement.
\end{itemize}

The common experimental settings are as follows: the number of episodes is set to 10000, with the observation missing rate varying from 0\% to 40\% in increments of 10\%. For each road network and demand level, we generate 100 traffic flow files. During training, each episode randomly samples one flow file, whereas evaluation is conducted over all 100 flow files. The missing observation mask is resampled at the beginning of each episode to cover diverse lane-level missing patterns. The discount factor $\gamma$ is set to 0.95. Each episode is set to simulate one hour of real time traffic. All agents execute actions every $\Delta t=10$ seconds, and there is no yellow phase between consecutive signal phases. Thus, the length $T$ of each episode is 360. All algorithms are optimized using the Adam~\cite{kingma2014adam} optimizer. The hyperparameters used for training the models are listed in Table~\ref{tab:hyperparameters}. For HALight, the latent dimension $d_h$ and the attention dimension $d_{a}$ are both set to 16, while the information dimension $d_{\text{info}}$ is set to 8. All experiments are conducted using the CityFlow~\cite{zhang2019cityflow} simulation platform on a Linux server equipped with an Intel(R) Xeon(R) Gold 6226R CPU and an NVIDIA GeForce RTX 3090 GPU.
\subsection{Metric}
We adopt four metrics to evaluate the agents' performance. All time-related metrics are measured in seconds.
\begin{itemize}
    \item \textbf{Average Travel Time (ATT)}: ATT measures the average time it takes for vehicles to travel from their origin to the destination. It helps evaluate how efficiently the traffic control system allows vehicles to complete their journey.

    \item \textbf{Average Queue Length (AQL)}: AQL quantifies the average number of vehicles waiting at each intersection throughout the episode. This is an important indicator of congestion. Formally, AQL is defined as the average queue length across all intersections and timesteps.

    \item \textbf{Average Waiting Time (AWT)}: AWT measures the average time that vehicles spend moving at speeds below 5 km/h during their journey. It provides insight into how much time vehicles spend in congested conditions.

    \item \textbf{Throughput (TP)}: TP measures the total number of vehicles that successfully complete their trips and reach their destinations during an episode. It is used to evaluate the overall traffic handling capacity. 
\end{itemize}

\begin{figure*}[t]
    \centering
    \includegraphics[width=\textwidth]{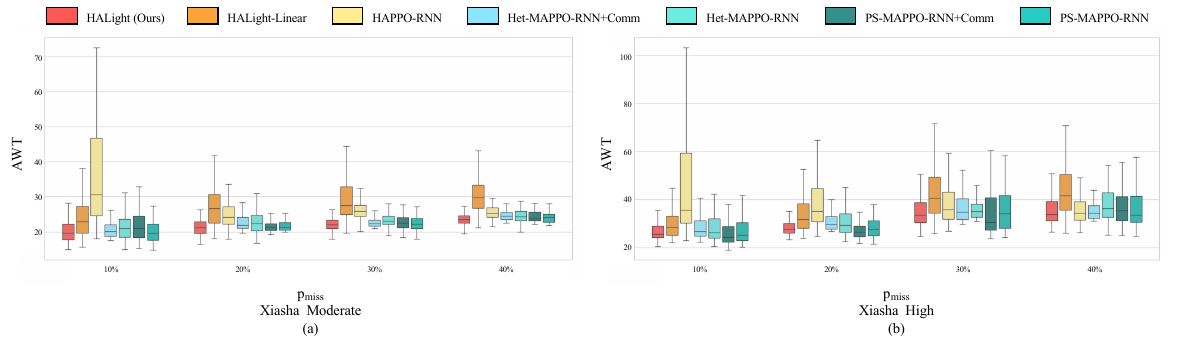}
    \caption{Ablation results of AWT on the Xiasha road network under different observation missing rates: (a) moderate demand and (b) high demand.}
    \label{fig:ablation}
\end{figure*}

\subsection{Performance Comparison}

The quantitative results are summarized in Tables~\ref{tab:Xiasha results} and~\ref{tab:Gudang results}, while the learning curves of the RL-based methods are shown in Fig.~\ref{fig:training_curves}. Overall, HALight achieves strong performance across both road networks, traffic demands, and observation missing rates, with its advantage becoming more evident when high traffic demand coincides with partial observability.

Under full observability ($p_{\text{miss}}=0\%$), both homogeneous and heterogeneous methods learn effective control policies. On the Gudang road network, PS-DynSTGAT achieves the best ATT and AQL among the RL-based methods under full observability, whereas heterogeneous methods generally perform better on the more structurally diverse Xiasha road network. As $p_{\text{miss}}$ increases, however, homogeneous baselines become substantially more sensitive to missing observations and even fail to converge in several settings. By comparison, HALight and Het-MAPPO-RNN remain considerably more stable. This suggests that intersection-specific observation and action modeling enhances adaptation to both observation missingness and structural heterogeneity, allowing each agent to learn a policy better suited to its local intersection characteristics.

\begin{table}[t]
\centering
\caption{Configuration of ablation variants. Homo and Hetero denote homogeneous and heterogeneous policy modeling.}
\label{tab:ablation_variants}
\scriptsize
\setlength{\tabcolsep}{4pt}

\resizebox{\linewidth}{!}{%
\begin{tabular}{lcccc}
\hline
\textbf{Variant} 
& \textbf{Policy Modeling} 
& \textbf{Comm} 
& \textbf{RNN} 
& \textbf{Sequential Update} \\
\hline
PS-MAPPO-RNN       & Homo   & $\times$      & $\checkmark$ & $\times$ \\
PS-MAPPO-RNN+Comm  & Homo   & $\checkmark$  & $\checkmark$ & $\times$ \\
Het-MAPPO-RNN      & Hetero & $\times$      & $\checkmark$ & $\times$ \\
Het-MAPPO-RNN+Comm & Hetero & $\checkmark$  & $\checkmark$ & $\times$ \\
HAPPO-RNN          & Hetero & $\times$      & $\checkmark$ & $\checkmark$ \\
HALight-Linear     & Hetero & $\checkmark$  & $\times$     & $\checkmark$ \\
HALight            & Hetero & $\checkmark$  & $\checkmark$ & $\checkmark$ \\
\hline
\end{tabular}%
}
\end{table}

Nevertheless, heterogeneous modeling alone is insufficient to ensure robustness under partial observability. HAPPO-RNN becomes unstable in several missing-observation scenarios; for example, on Gudang road network under high demand, its AQL increases from $2.72$ at $p_{\text{miss}}=0\%$ to $11.12$ at $40\%$. A plausible explanation is that incomplete observations increase policy uncertainty. This uncertainty may propagate through sequential policy updates as each agent accounts for previously updated policies. Het-MAPPO-RNN adopts independent policy updates and remains more stable in these settings. This indicates that the effectiveness of sequential updates can be significantly affected by partial observability and requires reliable complementary information.

HALight provides such complementary information through recurrent temporal modeling and one-hop neighborhood communication, and consequently achieves the most consistent performance under missing observations. Compared with Het-MAPPO-RNN, HALight generally obtains lower AQL together with improvements in ATT and AWT. On the Gudang road network under high demand, HALight reduces AQL by $6.38\%$, $2.82\%$, $7.94\%$, and $6.77\%$ relative to Het-MAPPO-RNN at $p_{\text{miss}}$ equals to $10\%$, $20\%$, $30\%$, and $40\%$. Notably, even on the highly heterogeneous Xiasha road network under high demand and severe observation missingness, HALight maintains competitive average performance despite increased variance. At $p_{\text{miss}}=40\%$, Fixed Time achieves a slightly higher TP than HALight, whereas HALight still achieves lower ATT, AQL, and AWT. Thus, HALight maintains comparable throughput while retaining advantages in the other evaluated metrics under this challenging setting. The learning curves in Fig.~\ref{fig:training_curves} further show that HALight generally maintains stable learning and high converged rewards as $p_{\text{miss}}$ increases. Unlike parameter-sharing methods, HALight learns intersection-specific policies without cross-agent experience sharing. This introduces a trade-off between experience utilization and policy specialization: the lack of cross-agent experience sharing reduces the benefit of collective experiences, especially in road networks where only a few intersections share similar structural characteristics. In contrast, independent policies allow agents to preserve intersection-specific behaviors and avoid policy averaging, leading to improved robustness under heterogeneous and partially observable environments.

\subsection{Ablation Study}

To evaluate the contribution of different components in HALight, we conduct ablation experiments on the Xiasha road network under moderate and high traffic demands, with $p_{\text{miss}}$ ranging from $10\%$ to $40\%$. As summarized in Table~\ref{tab:ablation_variants}, seven variants are constructed based on four design factors. ``Policy Modeling'' denotes whether homogeneous or heterogeneous policies are adopted, ``Comm'' indicates the use of one-hop neighborhood communication, ``RNN'' represents recurrent temporal modeling, and ``Sequential Update'' denotes whether sequential policy updates are employed. The corresponding AWT distributions are reported in Fig.~\ref{fig:ablation}.

\begin{itemize}

\item \textbf{Heterogeneous policy modeling.}
Compared with PS-MAPPO-RNN and PS-MAPPO-RNN+Comm, their heterogeneous counterparts generally achieve better performance, particularly under more challenging conditions. This indicates that intersection-specific policies better accommodate structural differences among intersections.

\item \textbf{One-hop neighborhood communication.}
The effect of communication can be evaluated by comparing PS-MAPPO-RNN with PS-MAPPO-RNN+Comm, Het-MAPPO-RNN with Het-MAPPO-RNN+Comm, and HAPPO-RNN with HALight. For both homogeneous and heterogeneous MAPPO variants, introducing communication generally improves performance or reduces performance variation under missing observations. The improvement is more evident when comparing HAPPO-RNN with HALight, where HALight achieves lower AWT and a more concentrated distribution in most settings. These results show that one-hop neighboring communication can alleviate the negative effects of incomplete local observations and improve robustness.


\item \textbf{Recurrent temporal modeling.}
HALight-Linear replaces the recurrent layer in HALight with a linear layer while retaining the other components. Compared with HALight-Linear, HALight achieves lower median AWT and more concentrated distributions, especially under high demand. This indicates that recurrent modeling enables agents to aggregate historical observations and capture temporal dependencies. It helps compensate for missing information and learn richer latent representations beyond the spatial context provided by communication.


\item \textbf{Sequential policy update.}
The effect of sequential policy updates can be examined by comparing HAPPO-RNN with Het-MAPPO-RNN, as well as HALight with Het-MAPPO-RNN+Comm. Without communication, HAPPO-RNN exhibits larger performance variations than Het-MAPPO-RNN in several settings, indicating that sequential updates can be sensitive to incomplete observations. With one-hop neighborhood communication, HALight generally achieves better and more stable performance than Het-MAPPO-RNN+Comm. These results indicate that, although sequential updates are theoretically analyzed under full observability, they can still effectively facilitate heterogeneous-agent coordination in partially observable scenarios when complemented with sufficient spatial and temporal information.

\end{itemize}

Taken together, the ablation results demonstrate that these designs jointly improve the robustness and effectiveness of HALight under lane-level missing observations.

\begin{figure}[t]
    \centering
    \includegraphics[width=\linewidth]{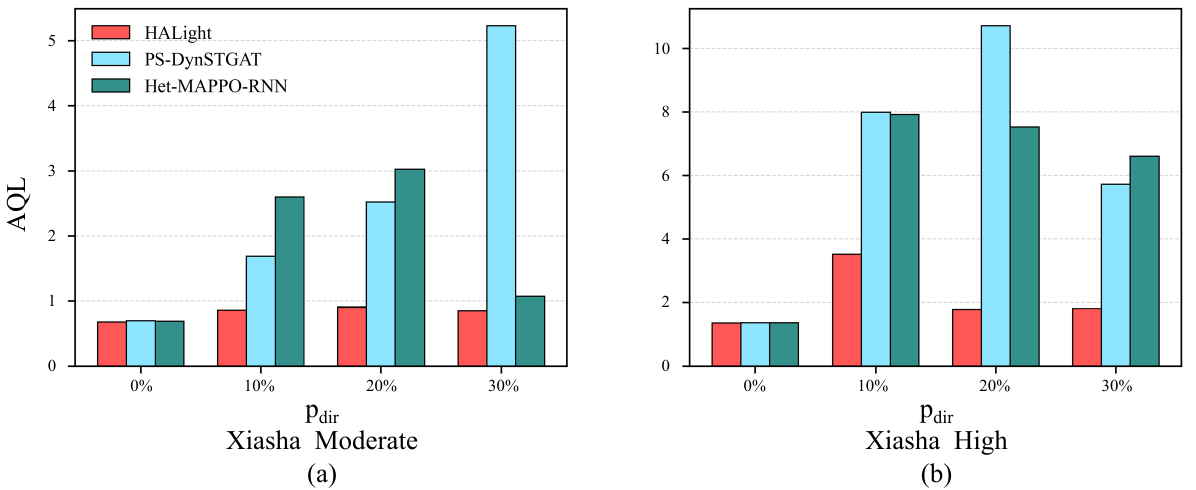}
    \caption{AQL performance under direction-level missing observations on the Xiasha road network: (a) moderate demand and (b) high demand.}
    \label{fig:direction_missing}
\end{figure}

\begin{figure}[t]
    \centering
    \includegraphics[width=\linewidth]{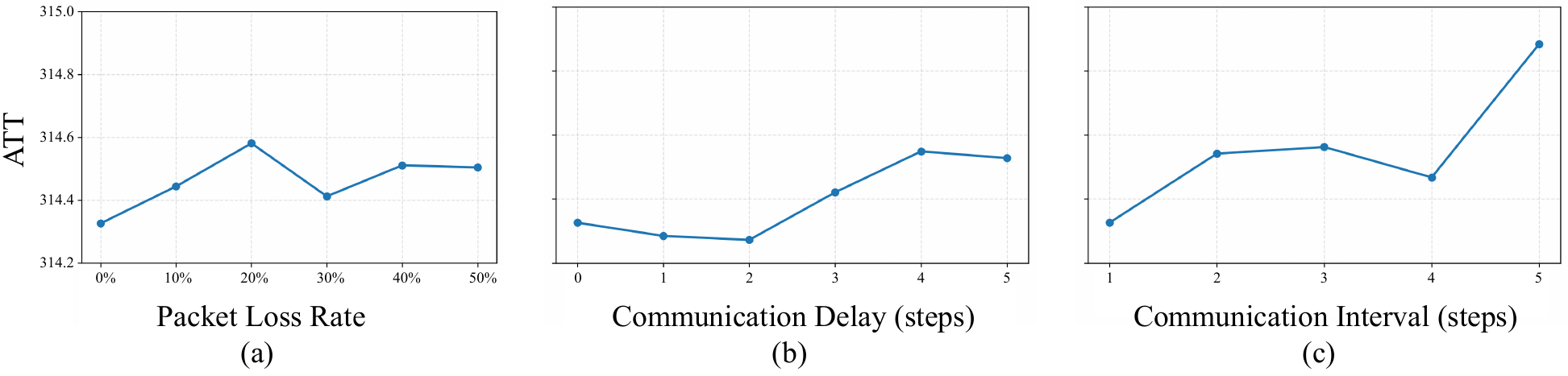}
    \caption{ATT of HALight under different communication conditions: (a) packet loss, (b) communication delay, and (c) communication interval.}
    \label{fig:communication_robustness}
\end{figure}

\subsection{Performance under Direction-Level Missing Observations}

To further evaluate performance under correlated observation missingness, we consider a direction-level missing setting on the Xiasha road network. Such grouped observation missingness is common in practice, where sensing failures or coverage limitations may simultaneously affect multiple lanes belonging to the same traffic direction. We introduce a direction-level missing probability $p_{\text{dir}}$. At each intersection, with probability $p_{\text{dir}}$, one traffic direction is randomly selected and the observations of all lanes in that direction are masked simultaneously. We compare HALight with PS-DynSTGAT and Het-MAPPO-RNN, which are the best-performing homogeneous and heterogeneous baselines, respectively. Both moderate and high traffic demands are considered, with $p_{\text{dir}}$ ranging from $0\%$ to $30\%$. The corresponding AQL results are shown in Fig.~\ref{fig:direction_missing}.

Under moderate demand, HALight maintains consistently lower AQL across different direction-level missing probabilities compared with PS-DynSTGAT and Het-MAPPO-RNN. The advantage of HALight becomes more evident under high demand, where the two baselines experience substantial performance degradation as $p_{\text{dir}}$ increases. In comparison, HALight achieves the best AQL performance across all direction-level observation missing scenarios.

These results show that HALight maintains strong performance under direction-level missing observations. By leveraging temporal information and one-hop neighborhood communication, HALight handles correlated missing observations across multiple lanes of the same direction, demonstrating adaptability to more structured missing patterns.

\subsection{Robustness to Communication Imperfections}

Reliable agent communication cannot always be guaranteed in practical deployments, as message transmission can be affected by bandwidth limitations, packet loss, and communication delays. To evaluate HALight under such non-ideal communication conditions, we introduce three types of communication perturbations during evaluation: packet loss, communication delay, and communication interval. For packet loss, the information received from the affected neighboring agent is replaced with an all-zero vector. HALight is trained under ideal communication conditions, whereas communication perturbations are introduced only during validation. We select the Gudang road network under high demand with $p_{\text{miss}}=40\%$ as the evaluation scenario. Specifically, the packet loss rate is varied from $0\%$ to $50\%$, the communication delay from 0 to 5 steps, and the communication interval from 1 to 5 steps. The corresponding ATT results are shown in Fig.~\ref{fig:communication_robustness}.

HALight maintains relatively stable performance across all three communication perturbations. With packet loss rates up to $50\%$, ATT remains close to the loss-free setting. Communication delays of up to 5 steps also cause only limited degradation. Increasing the communication interval has a modest effect in most cases, with a larger ATT increase observed at 5 steps. Across all settings, the ATT change remains below approximately $0.6$s compared with ideal communication.

These results demonstrate that HALight remains robust to communication imperfections after training, with only minor performance degradation under packet loss, communication delay, and reduced communication frequency.

\subsection{HALight Communication Efficiency Analysis}
In this subsection, we analyze the communication efficiency of HALight. The communication related computation at each timestep mainly consists of three components: Local Observation Encoding, Attentive Neighborhood Representation Aggregation, and Information Generation. At the system level, their per-timestep computational costs are $\mathcal{O}(n d_{o} d_{h}+|E| d_{\text{info}} d_{h})$, $\mathcal{O}(|E| d_{h} d_{a})$, and $\mathcal{O}(n d_{a} d_{\text{info}})$, respectively. Consequently, the overall communication related computation of HALight can be expressed as
\begin{equation}
    T=\mathcal{O}\big (n(d_{o} d_{h}+ d_{a}d_{\text {info }})+|E|(d_{\text {info }}d_{h}+d_{h} d_{a})\big ),
\end{equation}
where $d_o$ denotes the observation dimension, $n$ is the number of agents, and $|E|$ is the number of edges in the road network.

As indicated by the above expression, $T$ scales linearly with both $n$ and $|E|$, suggesting that HALight can be extended to large-scale road networks. Moreover, HALight requires only a single round of one-hop neighborhood aggregation per decision, avoiding the latency and computation introduced by multi-round message passing. Empirically, we find that small dimensions $d_h$, $d_a$, and $d_{\text{info}}$ remain sufficient to achieve substantial performance gains even when they are much smaller than the hidden width of the policy network, indicating that lightweight communication can effectively enhance coordination under lane-level missing observations.

\section{Conclusion}
\label{sec:conclusion}

In this paper, we studied TSC in realistic settings where intersections are structurally heterogeneous and lane-level observations can be partially missing. We proposed HALight, a communication-enhanced heterogeneous MARL framework. HALight supports intersection-specific observation/action modeling and integrates lightweight one-hop neighborhood communication with temporal modeling, while employing a sequential policy update scheme to stabilize heterogeneous agent learning. Extensive experiments on two real-world road networks, covering different network scales, traffic demands, and observation missing rates, demonstrate HALight's effectiveness. Our communication efficiency analysis also shows that the proposed communication design scales favorably with network size.

Despite these promising results, several directions remain for future work. First, it is worthwhile to evaluate the proposed framework under more realistic traffic conditions, including richer phase plans and diverse sensing/communication failure patterns. Second, further validation on larger-scale road networks is needed, and exploring adaptive communication beyond one-hop neighborhoods may improve performance when traffic interactions extend over longer distances.

\appendices
\section{Convergence Analysis of HALight under Full Observability}
\label{app:convergence}

We analyze the convergence property of HALight under full observability. In this case, each agent conditions its policy on the Markovian state \(s_t\in\mathcal S\), and the joint policy is factorized as 
\(\boldsymbol{\pi}_{\theta}(\bm a_t|s_t)=\prod_{i=1}^{n}\pi_{\theta^i}^{i}(a_t^i|s_t)\), where \(\bm a_t=(a_t^1,a_t^2,\dots,a_t^n)\in\mathcal A\). 
The communication encoder, attention aggregation, information decoder, and policy head are treated as differentiable components of the individual policies and do not alter this factorization. At iteration \(k\), HALight updates the actors sequentially according to a random permutation \((i_1,i_2,\dots,i_n)\), with each update weighted by the correction factor \(M_{i_{1:m}}\). Therefore, under full observability, HALight follows the HAPPO-style optimization structure with factorized heterogeneous policies, sequential agent-wise updates, and correction for previously updated agents.

\textbf{Proposition.} Under the fully observable cooperative Markov game setting and the standard assumptions of HAPPO, HALight satisfies the monotonic improvement property, i.e.,
\[
J(\boldsymbol{\pi}_{\theta_{k+1}})
\geq
J(\boldsymbol{\pi}_{\theta_k}), \quad \forall k.
\]
Furthermore, if every agent permutation has a fixed non-zero probability of being sampled, the limit points of the generated joint-policy sequence are Nash equilibria.

\textit{Proof.} Under full observability, the HALight actor of each agent defines a valid individual policy \(\pi_{\theta^i}^{i}(a^i|s)\). 
Since the communication and attention modules only affect the internal representation of \(\pi_{\theta^i}^{i}\), the joint policy still factorizes as \(\boldsymbol{\pi}_{\theta}(\bm a|s)=\prod_{i=1}^{n}\pi_{\theta^i}^{i}(a^i|s)\).

For any permutation \((i_1,i_2,\dots,i_n)\), the multi-agent advantage decomposition lemma gives
\[
A_{\boldsymbol{\pi}}(s,\bm a)
=
\sum_{m=1}^{n}
A_{\boldsymbol{\pi}}^{i_m}
\left(s,\bm a^{i_{1:m-1}},a^{i_m}\right).
\]
Thus, the improvement of the joint policy can be evaluated through a sequence of agent-wise advantage terms. 
For agent \(i_m\), HALight uses \(M_{i_{1:m}}\) to account for the policies of agents \(i_1,\dots,i_{m-1}\) that have already been updated. 
Therefore, the update objective of \(i_m\) is equivalent to optimizing its contribution to the decomposed joint advantage under the updated previous agents.

Let \(\boldsymbol{\pi}_{\theta_k}\) and \(\boldsymbol{\pi}_{\theta_{k+1}}\) denote the joint policies before and after one sequential update. 
Following the HAPPO policy improvement bound, the return difference is lower-bounded by the sum of the sequential improvement terms:
\[
J(\boldsymbol{\pi}_{\theta_{k+1}})-J(\boldsymbol{\pi}_{\theta_k})
\geq
\sum_{m=1}^{n}
\Delta_{i_m},
\]
where \(\Delta_{i_m}\) denotes the improvement term obtained by updating agent \(i_m\) with the correction factor \(M_{i_{1:m}}\). 
Since agent \(i_m\) can always choose the old policy \(\pi_{\theta_k^{i_m}}^{i_m}\), which yields zero improvement, maximizing the HAPPO-style objective gives \(\Delta_{i_m}\geq 0\) for every \(m\). 
Therefore,
\[
J(\boldsymbol{\pi}_{\theta_{k+1}})-J(\boldsymbol{\pi}_{\theta_k})
\geq
0,
\]
which proves the monotonic improvement property.

Moreover, the random update order in HALight assigns non-zero probability to every permutation. 
This satisfies the update-order condition in HAPPO convergence analysis. 
Hence, under the same full observation assumptions, the limit points of HALight's joint policy sequence are Nash equilibria.
\(\hfill\Box\)


%





\ifCLASSOPTIONcaptionsoff
  \newpage
\fi



%
\bibliographystyle{IEEEtran}
\bibliography{references}
%







\end{document}